\documentclass[journal]{IEEEtran}
\ifCLASSINFOpdf
\else
\fi

\usepackage{amsmath}
\usepackage{hyperref}
\usepackage{cite}
\usepackage{color}
\usepackage{booktabs}
\usepackage{threeparttable}
\usepackage{graphicx}
\usepackage{multirow}
\usepackage{footnote}
\usepackage{threeparttable}
\usepackage{longtable}
\usepackage{makecell}
\usepackage[noend]{algpseudocode}
\usepackage{algorithmicx,algorithm}
\usepackage{float}
\usepackage{amssymb}

\begin{document}
%
\title{UFA-FUSE: A novel deep supervised and hybrid model for multi-focus image fusion}
%
%
%

\author{Yongsheng~Zang,
        Dongming~Zhou,
        Changcheng~Wang,
        Rencan~Nie,
        and~Yanbu~Guo
\thanks{This work was supported by the National Natural Science Foundation
of China under Grants 62066047, 61966037, 61365001 and 61463052,
and Yunnan Province University Key Laboratory Construction Plan
Funding, China. (Corresponding author: Dongming Zhou.)}

\thanks{Y. Zang is with the School of Information Science and Engineering, Yunnan University, Kunming 650091, China (email: 12019202218@mail.ynu.edu.cn). }
\thanks{D. Zhou is with the School of Information Science and Engineering, Yunnan University, Kunming 650091, China (e-mail:zhoudm@ynu.edu.cn)}
\thanks{C. Wang is  with the School of Information Science and Engineering, Yunnan University, Kunming 650091, China (email: changcheng@mail.ynu.edu.cn).}
\thanks{R. Nie is with the School of Information Science and Engineering, Yunnan University, Kunming 650091, China (e-mail:rcnie@ynu.edu.cn).}
\thanks{Y. Guo is with the School of Information Science and Engineering, Yunnan University, Kunming 650091, China (email:guoyanbu@gmail.com).}}
\markboth{}%
{Shell \MakeLowercase{\textit{.}}: Bare Demo of IEEEtran.cls for IEEE Journals}
%



\maketitle

\begin{abstract}
Traditional and deep learning-based fusion methods generated the intermediate decision map to obtain the fusion image through a series of post-processing procedures. However, the fusion results generated by these methods are easy to lose some source image details or results in artifacts. Inspired by the image reconstruction techniques based on deep learning, we propose a multi-focus image fusion network framework without any post-processing to solve these problems in the end-to-end and supervised learning ways. To sufficiently train the fusion model, we have generated a large-scale multi-focus image dataset with ground-truth fusion images. What’s more, to obtain a more informative fusion image, we further designed a novel fusion strategy based on unity fusion attention, which is composed of a channel attention module and a spatial attention module. Specifically, the proposed fusion approach mainly comprises three key components: feature extraction, feature fusion and image reconstruction. We first utilize seven convolutional blocks to extract the image features from source images. Then, the extracted convolutional features are fused by the proposed fusion strategy in the feature fusion layer. Finally, the fused image features are reconstructed by four convolutional blocks. Experimental results demonstrate that the proposed approach for multi-focus image fusion achieves remarkable fusion performance and superior time efficiency compared to 19 state-of-the-art fusion methods.
\end{abstract}


%
\IEEEpeerreviewmaketitle

\section{Introduction}
%
%
%
%
\IEEEPARstart {I}{m}age fusion\cite{vishwakarma2018image} is an effective approach to extend the measurement of optical lenses to integrate two or more input images to produce a more informative fused image. Image fusion includes several fusion subareas such as multi-focus image fusion, infrared and visible image fusion{{\cite{9345717},\cite{9187663}}} and medical image fusion{\cite{9305718},\cite{8385209}}. Specifically, multi-focus image fusion aims to get clear images that obtain all objects of a scene in focus. Multi-focus image fusion has become more closely related to our daily lives, while playing a crucial role in biological, industrial and medical fields. For example, multi-focus image fusion is used to fuse microscopic cervical cell images for disease diagnosis\cite{2019Efficient}, and can also be applied to structural measurement of nonwoven fabrics\cite{2019Structural}. Also, multi-focus image fusion is a guarantee for the precise acquisition of pap smear images\cite{2018Region}.

Generally speaking, there exist three categories of multi-focus image fusion techniques according to imaging principles. The fusion techniques are broadly divided into transform domain techniques \cite {stathaki2011image}, spatial domain techniques and deep learning-based methods. The transform domain techniques are prevailing in the first phase of image fusion. The image algorithms based on transform domain share a universal three main steps to fuse image, which are generally summarized as decomposition, fusion and reconstruction. Some representative examples include pyramid-based and wavelet-based methods, such as the morphological pyramid-based method \cite {toet1989morphological}, the discrete wavelet transform-based method\cite{li1995multisensor}, the stationary wavelet transform-based method\cite{2018A}, the dual-tree complex wavelet transform-based method \cite{lewis2007pixel}, the adjustable non-subsampled shearlet transform \cite{vishwakarma2018image} and the non-subsampled contourlet transform-based method \cite{zhang2009multifocus}.

The spatial domain methods do not require the source image to be converted to another feature domain like the transform domain methods. Spatial domain-based methods operate on pixels or regions directly and use linear or nonlinear methods to fuse the multi-focus images. Spatial domain-based methods can be roughly divided into three categories \cite{bai2015quadtree}, which often include pixel-based \cite{li2013image}, block-based \cite{bai2015quadtree} and region-based \cite{li2008multifocus} algorithms. The block-based algorithms decompose the input images into blocks with equal size according to block-based strategy. Then the decomposed blocks are fused by the designed activity level measurement. However, the selected block-size has a great impact on the quality of the fused image, the unsuitable size of block restricts the performance of these block-based algorithms. The region-based algorithms firstly use image segmentation to split up the input images into regions. Then, the clarities of corresponding regions are measured and image fusion is performed by integrating the sharply focused regions. Clearly, the accuracy of image segmentation greatly affects the efficiency of the region-based algorithms. Different from the fusion algorithms mentioned above, pixel-based algorithms directly use activity level measurement strategy to generate the decision map for fusion. Some novel pixel-based spatial domain methods have been proposed, such as multi-scale weighted gradient (MWGF) \cite{zhou2014multi}, guided filtering (GFF) \cite{li2013image} and dense SIFT \cite{liu2015multi}.

Although the traditional fusion methods have indicated some significant advantages, these fusion methods still exist some weaknesses as follows: (i) The fusion methods rely highly on activity level measurement strategy and transform domain, which leads to poor robustness of these fusion methods. (ii) These fusion methods have poor fusion performance when the input images are complex. To ameliorate these shortcomings, a series of deep learning-based fusion algorithms have been proposed and obtained good results. For example, Liu \emph{et al}. \cite{liu2017multi} proposed convolutional neural network (CNN)-based fusion model which can recognize whether the image area is in-focus to generate the decision map for image fusion. Based on this, Tang \emph{et al}. \cite{tang2018pixel} optimized the structure of CNN from the perspective of pixel to obtain the decision map. Besides, Ma \emph{et al}. \cite{ma2020sesf} proposed an unsupervised deep learning fusion framework, SESF, to generate a well decision map for image fusion. It is no doubt that these fusion models reached a better performance compared to traditional methods because of the strong and stabilized representation capability. However, the fusion image generated by these methods heavily rely on the decision map refined by some post-processing procedures, which is hard to obtain the fusion image with high quality. Therefore, other fusion models, such as DeepFuse \cite{prabhakar2017deepfuse}, DenseFuse \cite{li2018densefuse} and IFCNN \cite{zhang2020ifcnn} directly fused the deep image features rather than generating the decision map to reconstruct the fusion image. Although the above methods have achieved relatively good performance in image fusion, the new problems which are lacking of rich image details yielded with simple fusion strategy remain unsolved in these image fusion methods. The conclusion of the existing SOTA methods is shown in Table I.\\
\begin{table*}[ht]
	\centering
	\caption{The conclusion of the existing SOTA methods}
	\setlength{\tabcolsep}{4mm}{
		\begin{tabular}{|c|c|l|l|}
			\hline
			Classes                                                                      & Methods                                                                                              & \multicolumn{1}{c|}{Advantages}                                                                                                                                                                                                                                                                                                        & \multicolumn{1}{c|}{Disadvantages}                                                                                                                                                                                                                                                                                                                                      \\ \hline
			\begin{tabular}[c]{@{}c@{}}Transform\\ domain-based\\ methods\end{tabular} & \begin{tabular}[c]{@{}c@{}}MP\cite{toet1989morphological}\\ DWT\cite{li1995multisensor}\\CVT\cite{nencini2007remote} \\ SWT\cite{2018A}\\ DTCWT\cite{lewis2007pixel}\end{tabular}                                 & \begin{tabular}[c]{@{}l@{}}1.These methods have superior expansion \\ performance and can be used to fuse \\ different types of images.\\ 2. These methods can utilize local and\\ multi-resolution image features more\\ effectively, and the fusion results are\\ richer in details.\end{tabular}                                    & \begin{tabular}[c]{@{}l@{}}1. The fusion results of these methods\\ are susceptible to artifacts due to\\ the halo and blurring effect.\\ 2. The performance of these methods is\\ limited by the finite choice of fusion\\ rules and the complexity of manual \\ design, and salient image features are\\ easily lost during the fusion process.\end{tabular}          \\ \hline
			\begin{tabular}[c]{@{}c@{}}Spatial \\ domain-based \\ methods\end{tabular} & \begin{tabular}[c]{@{}c@{}}IMF\cite{chen2017robust}\\  GFF\cite{li2013image}\\ SF\cite{li2008multifocus}\\ MWGF\cite{zhou2014multi}\\ Dense SIFT\cite{liu2015multi}\end{tabular}                             & \begin{tabular}[c]{@{}l@{}}1. The performance of these algorithms is\\  robust and not easily affected by noise \\ and mis-alignment.\\ 2. These methods do notneed to transform\\ the image to some other feature domain\\ to achieve fusion, which can effectively\\ reduce blurring effects and spatial\\  distortion.\end{tabular} & \begin{tabular}[c]{@{}l@{}}1. these methods tend to lose the \\ original image information due to the\\ inaccurate fusion decision map.\\ 2. These algorithms usually contain\\ more complex processing procedures and\\ have high computational overhead. In\\ addition, the algorithms are\\ inefficient and time-consuming when\\ fusing source images.\end{tabular} \\ \hline
			\begin{tabular}[c]{@{}c@{}}Deep \\ learning-based\\  methods\end{tabular}  & \begin{tabular}[c]{@{}c@{}}CNN\cite{liu2017multi}\\ p-CNN\cite{tang2018pixel}\\ SESF\cite{ma2020sesf}\\ IFCNN\cite{zhang2020ifcnn}\\ DenseFuse\cite{li2018densefuse}\\ DeepFuse\cite{prabhakar2017deepfuse}\\ DRPL\cite{9020016}\\ MFF\cite{ZHANG202140}\end{tabular} & \begin{tabular}[c]{@{}l@{}}1. These deep learning-based methods can\\  automatically learn to extract features\\  without the need to design complex \\ extraction approaches.\\ 2. Deep learning based methods can\\  preserve more informative image features \\ than complex handcrafted feature \\ extraction operators. \end{tabular}                                                          & \begin{tabular}[c]{@{}l@{}}1.Some deep learning-based methods\\ rely on generating intermediate\\ decision maps toachieve image fusion,\\ and the generated fused images are \\ prone to artifacts.\\ 2.The fusion strategy is too common to\\ effectively achieve the fusion of deep\\  features.\end{tabular}                                                         \\ \hline
		\end{tabular}
		\label{tbl:table-example}}
\end{table*}
\indent To address these problems mentioned above, in this paper, A novel fusion model is proposed for multi-focus images. The fusion model can be trained in end-to-end manner without any post-processing procedures, and directly obtain the fusion image without generating the intermediate decision map. To effectively train the fusion model, we have created a large-scale multi-focus image dataset with ground-truth fusion image. Moreover, to obtain a high quality of fusion image, we design a novel fusion strategy based on unity fusion attention mechanism. Specifically, the source images were firstly extracted in feature extraction module by seven stacked convolutional blocks. Then the extracted image features are fused by the proposed fusion strategy. Finally, the fused image features were reconstructed by four stacked convolutional blocks to generate the fusion image. Furthermore, we conducted extensive experiments to evaluate the performance of our proposed fusion model. The results of qualitative evaluation and quantitative evaluation show that our fusion model achieves the remarkable performance, which is superior to the state-of-the-art multi-focus image fusion algorithms. Overall, the main contributions of our work for multi-focus image fusion are fourfold:

\begin{itemize}
	\item A large-scale multi-focus image dataset is generated for multi-focus image fusion. The generated multi-focus images in data sets are partially-focused and have different levels of blur, which is suitable for training the fusion model. Moreover, we introduced how to generate the large-scale multi-focus image dataset, which is of great reference significance for making the dataset.
\end{itemize}

\begin{itemize}
	\item A novel and appropriate fusion strategy is designed. The proposed fusion strategy based on proposed unity fusion attention can effectively integrate the extracted image features and meanwhile is more flexible than other simple fusion strategies.
\end{itemize}

\begin{itemize}
	\item An end-to-end fusion framework without any post-processing procedures is proposed. So, all the parameters in the fusion model can be jointly optimized and the fusion network can directly input the fusion image without generating the intermediate decision map. In addition, the fusion framework is very efficient and fast in processing images, substantially ahead of other comparison algorithms. 
\end{itemize}

\begin{itemize}
	\item Some expanded experiments on the infrared and visible image dataset and medical image dataset are conducted. Furtherly, the potential applications of our proposed fusion model for other fusion tasks are presented.
\end{itemize}

The remainder of our paper is structured as follows: we briefly review related works about fusion algorithms based on deep learning in Section II. Then Section III describes the proposed fusion framework and introduces the generated large-scale multi-focus image dataset in detail. In Section IV, we present the extensive experimental results and the expanded experiments. We also discuss the application of the proposed framework and the prospects for future research work in Section V. Finally, the conclusion is drawn in Section V. The codes of our proposed UFA-Fusion are accessible on https://github.com/ReckonerInheritor/UFA-Fusion.

\section{Related Works}
With the development of deep learning in recent years, a series of fusion algorithms based on deep learning have been proposed and also obtained remarkable results in computer image processing \cite{2020Brain} and biomedicine \cite{guo2020deepanf}. Moreover, various-deep learning-based methods have shown their own unique theories and advantages for image fusion. Consequently, we briefly introduce several representative image fusion methods in this section. Then we present the application of attention mechanism in various image fusion tasks.

\subsection{Deep learning-Based Algorithms}
As is well-known, the first deep learning-based algorithm is CNN-based fusion network which was proposed by Liu \emph{et al}. in 2017 \cite{liu2017multi}. The CNN-based network treated the image fusion task as image classification task, the image patches with different blur versions are fed into the fusion network to predict the focus map. Another CNN-based algorithm [13] proposed by Tang \emph{et al}. used the CNN model to learn an effective focus-measure from the perspective of image pixel, and then compared the value of focus-measures to generate the decision map. Similar to these two CNN-based fusion methods, SESF \cite{ma2020sesf} used an encoder network to extract image features, and then calculated the spatial frequency of image features to obtain the decision map. Afterward, the above all three fusion algorithms adopted a series of post-processing procedures, such as morphological operations and small block filter, to refine the generated decision map. The quality of fusion image depends heavily on how well the decision map is refined. Furthermore, image detail information is inevitably lost during post-processing, resulting in the artifacts in fusion image. 

What’s more, there are also some deep learning-based networks applied to image fusion tasks. Ram \emph{et al}. \cite{prabhakar2017deepfuse} proposed a deep learning-based fusion network, DeepFuse, for fusing the exposure images. In DeepFuse, two pre-fusion convolutional layers which that shared the same weights were used to extract low-level image features, and then the features are fused in a single merge layer by an elementwise-add fusion strategy. Finally, the fused image was reconstructed by three convolutional layers. Inspired by DeepFuse, Li \emph{et al}. \cite{liu2017multi} proposed an unsupervised encoder-decoder network, DenseFuse, for the infrared and visible fusion problem. DenseFuse consists of three parts: firstly, the encoder network was used to extract the image features, and fused them in a fusion layer by an L1-norm fusion strategy or  elementwise-add fusion strategy, then reconstructed by the decoder network. These two fusion models do not require the post-processing procedures to optimize the intermediate decision map. Consequently, these two fusion models can also fuse other types of images such as multi-focus images and medical images. Similar to DeepFuse and DenseFuse, Zhang \emph{et al}. \cite{zhang2020ifcnn} proposed the IFCNN which is a general fusion network based on CNN without any post-processing procedures. In IFCNN, the pre-trained convolutional layer of ResNet101 was firstly used to extract the image features, and then the image features were fused by different fusion rules according to the type of input image. The fusion rules include elementwise-maximum, elementwise-sum, and elementwise-mean. Finally, the fusion image was reconstructed by two convolutional layers. All these three fusion models do not require the post-processing procedures to produce fusion images and their models can be fully optimized for image fusion. However, there is a problem in these methods, namely, these methods adopted the fusion strategy based on simple mathematical operations, which ignored the deep image details in the process of feature fusion.
\begin{figure*}
	\centering
	\includegraphics[width=0.90\textwidth]{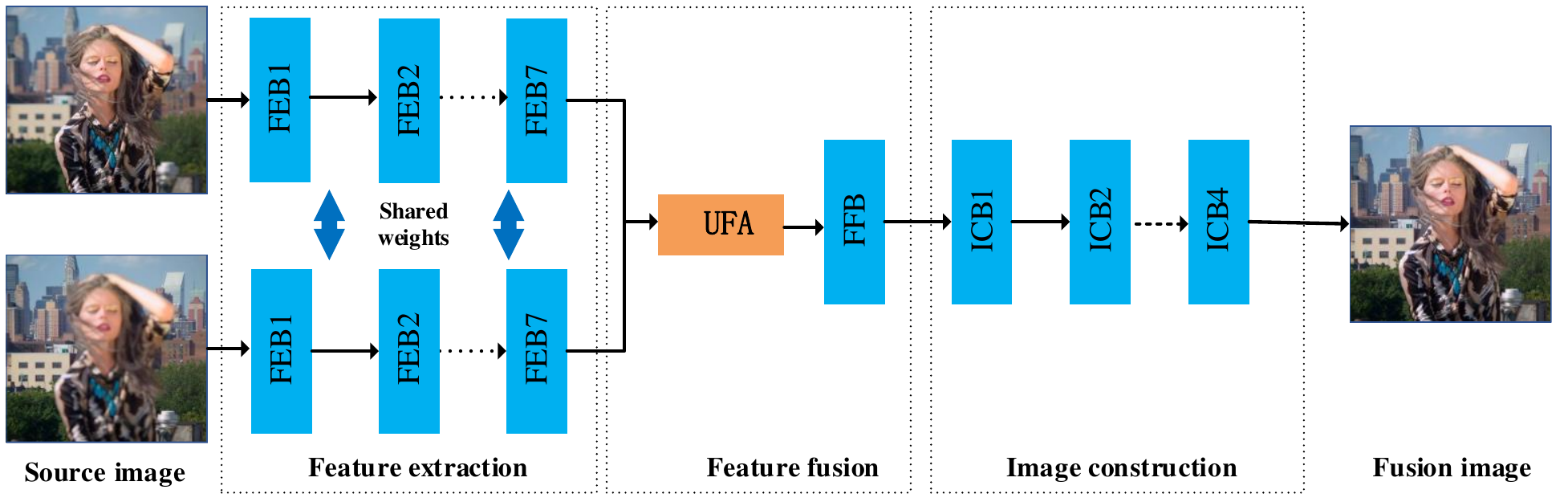}
	\vspace{-0.2 cm}
	\caption{The demonstration of the image fusion framework.}
	\vspace{-0.5 cm}
\end{figure*}

\subsection{Attention Mechanism}
Attention mechanism has been widely applied in computer vision tasks since the SENet \cite{hu2018squeeze} was proposed by Hu \emph{et al}. in 2017. It is proved that the attention mechanism was useful for deep learning as an effective component. In SENet, it firstly introduced channel-wise attention which calculated the correlation between different channels by global average pooling and fully connection layers. Since it is a simple and ‘plug and play’ component, it is also gradually applied in image fusion tasks. In SESF, the SE module was inserted to the encoder network to enhance the feature extraction capabilities of the network. Li \emph{et al}. \cite{9216075} introduced the attention module into the infrared and visible image fusion framework to explore multi-scale features and force the fusion framework to pay more attention to the discriminative regions. In NestFuse \cite{9127964}, Li \emph{et al}. proposed the cascaded spatial attention and channel attention as the fusion strategy to fuse the deep features in the infrared and visible image fusion framework. In GEU-Net\cite{9242278}, Xiao \emph{et al}. construct attention-based upsampling modules to efficiently extract and exploit global semantic and edge information to achieve superior fusion performance. What’s more, based on SENet, Sanghyun Woo \emph{et al}. proposed the convolutional block attention module (CBAM) \cite{woo2018cbam}. The module calculated the attention map along separate channels and spatial dimensions. Although CBAM is an effective and lightweight module, it is rarely seen to be applied in image fusion tasks. In fact, CBAM is suitable for application in single-image processing task and can improve the network’s feature representation capability. As we all know, multi-focus image fusion aims to integrate as many clear image features as possible from source images. Therefore, inspired by CBAM, we want to design a fusion strategy which can focus image features from channel and spatial these two dimensions. Based on this, this fusion strategy can preserve more image details compared to other simple fusion strategy based elementwise operations.

\section{The proposed fusion method}
In this section, we will amply introduce the proposed fusion network. Firstly, the fusion framework is presented in Section A. Then, we introduce the details of our proposed fusion strategy based on a novel attention model. Finally, we present a new way to generate a large-scale multi-focus image dataset. 
\subsection{Image fusion framework}
As shown in Fig. 1, our proposed fusion framework contains three main portions: feature extraction, feature fusion and image construction. Firstly, we adopt seven feature extraction blocks (FEB) to extract image features from multi-focus source images in the feature extraction module, and each of the FEB consists of a convolutional layer with kernel size of 3×3 and leakyRelu activation function. For each convolutional feature of input source images, our proposed feature fusion module is utilized to fuse the resulting features. Finally, the image construction module, which is composed of four image reconstruction blocks (ICB), is used to reconstruct the fused image features to generate the fusion image. The network settings of the fusion framework are shown in detail in Table II. Furthermore, to obtain a more accurate fusion image, we use the loss function $L$ to train our fusion network in the training phase.

\begin{table}[t]
	\centering
	\caption{The network settings of the fusion framework}
	\setlength{\tabcolsep}{0.4mm}{
		\begin{tabular}{|c|c|c|c|c|c|c|}
			\hline
			& Layer      & Size & Stride & \begin{tabular}[c]{@{}c@{}}Channel\\ (input)\end{tabular} & \begin{tabular}[c]{@{}c@{}}Channel\\ (output)\end{tabular} & Activation \\ \hline
			\multirow{7}{*}{\begin{tabular}[c]{@{}c@{}}Feature\\ \\ extraciton\end{tabular}} & Conv(FEB1) & 3    & 1      & 3                                                          & 64                                                         & LeakyRelu  \\ \cline{2-7} 
			& Conv(FEB2) & 3    & 1      & 64                                                         & 64                                                         & LeakyRelu  \\ \cline{2-7} 
			& Conv(FEB3) & 3    & 1      & 64                                                         & 64                                                         & LeakyRelu  \\ \cline{2-7} 
			& Conv(FEB4) & 3    & 1      & 64                                                         & 64                                                         & LeakyRelu  \\ \cline{2-7} 
			& Conv(FEB5) & 3    & 1      & 64                                                         & 64                                                         & LeakyRelu  \\ \cline{2-7} 
			& Conv(FEB6) & 3    & 1      & 64                                                         & 64                                                         & LeakyRelu  \\ \cline{2-7} 
			& Conv(FEB7) & 3    & 1      & 64                                                         & 64                                                         & LeakyRelu  \\ \hline
			\multirow{2}{*}{\begin{tabular}[c]{@{}c@{}}Feature\\ fusion\end{tabular}}        & Conv($f^{7 \times 7}$)    & 7    & 1      & 2                                                          & 1                                                          & Null       \\ \cline{2-7} 
		
			& Conv(FFB)  & 1    & 1      & 128                                                        & 64                                                         & Null       \\ \hline
			\multirow{4}{*}{\begin{tabular}[c]{@{}c@{}}Image\\ constrction\end{tabular}}     & Conv(ICB1) & 3    & 1      & 64                                                         & 64                                                         & LeakyRelu  \\ \cline{2-7} 
			& Conv(ICB2) & 3    & 1      & 64                                                         & 64                                                         & LeakyRelu  \\ \cline{2-7} 
			& Conv(ICB3) & 3    & 1      & 64                                                         & 64                                                         & LeakyRelu  \\ \cline{2-7} 
			& Conv(ICB4) & 3    & 1      & 64                                                         & 3                                                         & Sigmod     \\ \hline
		\end{tabular}
		\label{tbl:table-example}}
\end{table}

\begin{figure*}
	\centering
	\includegraphics[width=0.90\textwidth]{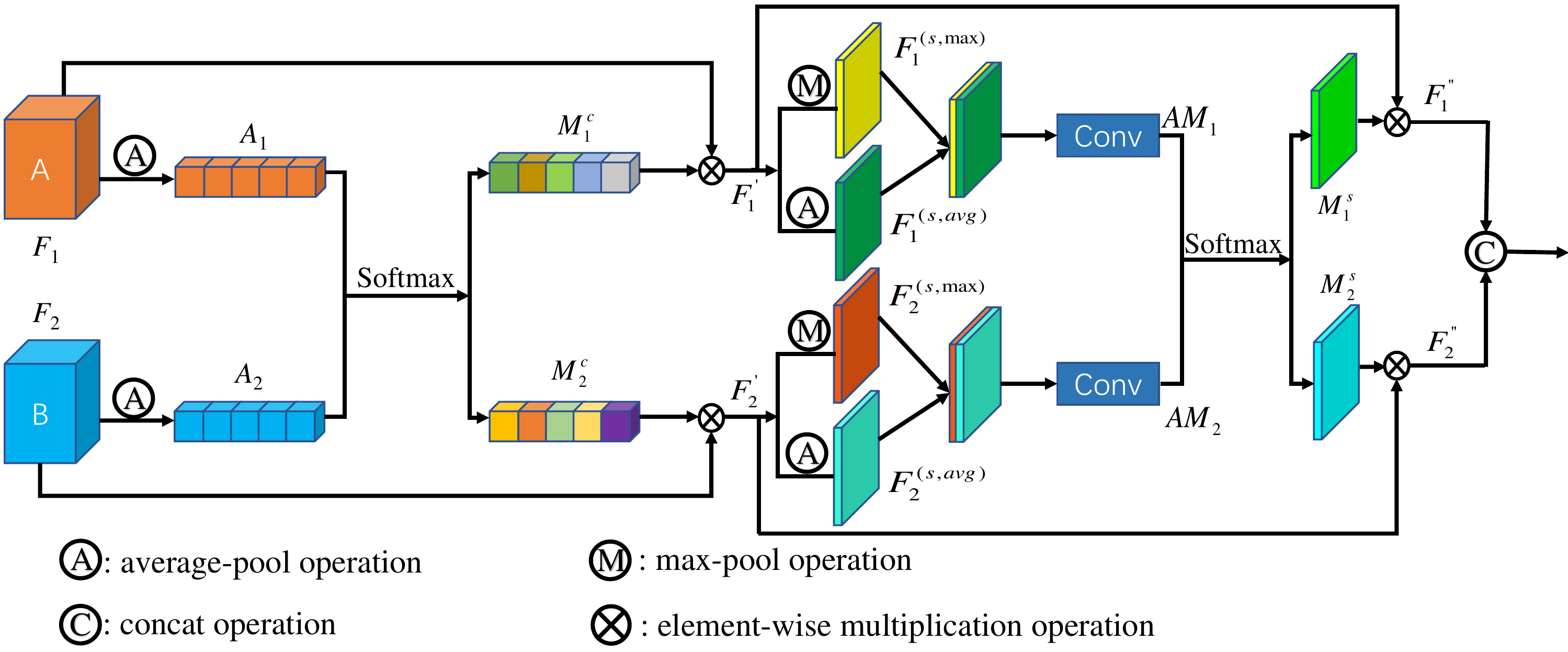}
	\vspace{-0.2cm}
	\caption{Diagram of the unity fusion attention module.}
	\vspace{-0.5cm}
\end{figure*}
The loss function $L$ is defined as follows,
\begin{equation}
	L=\lambda L_{1}+(1-\lambda) L_{SSIM},
\end{equation}
 where $L_{1}$ and $L_{SSIM}$ denote the $L_{1}$-norm loss function and structure similarity (\emph{SSIM}) loss function, and the $\lambda$ is the trade-off parameters between $L_{1}$ and $L_{SSIM}$.
 
The definition of $L_{1}$ loss function is described as below,
\begin{equation}
	L_{1}=|O-I|_{1},
\end{equation}
where ${I}$ and ${O}$ denote the predicted fusion image and ground-truth fusion image, respectively.

The $L_{SSIM}$ loss function is calculated as below,
\begin{equation}
	{SSIM}(O, I)=\frac{\left(2 \mu_{O} \mu_{I}+C_{1}\right)\left(\sigma_{O I}+C_{2}\right)}{\left(\mu_{O}^{2}+\mu_{I}^{2}+C_{1}\right)\left(\mu_{O}^{2}+\mu_{I}^{2}+C_{2}\right)},
\end{equation}
\begin{equation}
	L_{SSIM}=1-{SSIM}(O, I),
\end{equation}
where  ${SSIM} (*)$ denotes the structural similarity operation, which represents the structural similarity between the predicted fusion image and ground-truth fusion image. $\mu _{O}$ and $\mu _{I}$ indicate the mean value of $O$ and $I$, $\sigma _{O}$ and $\sigma _{I}$ indicate the standard deviation of $O$ and $I$, respectively. $\sigma _{OI}$ denotes the covariance between $O$ and $I$. $C_{1}$ and $C_{2}$ are constants to avoid the case where the denominator is 0. The parameter $\lambda$ was set to 0.2 by experimental and reference a priori empirical values\cite{7797130}. A detailed analysis of the values of the parameter $\lambda$ is presented in the ablation study given in Section IV-B.
\subsection{Fusion strategy}
According to previous works \cite{prabhakar2017deepfuse,li2018densefuse,zhang2020ifcnn}, we can know that most fusion strategies are based on straightforward fusion operators, such as elementwise-add, elementwise-mean, elementwise-maximum, and elementwise-sum. These simple fusion strategies fuse all extracted image features equally, which causes the loss of informative image features. This is why the fusion algorithms with simple fusion strategies are hard to recover a high visual quality fusion image to some extent.

As we have learned, the application of attention mechanism in infrared and visible image fusion task\cite{9216075,9127964,9242278} and inspired by CBAM\cite{woo2018cbam}, we further propose a novel fusion strategy based on attention mechanism in the process of feature fusion. The feature weights of the attention-based feature fusion structure are learned adaptively by the attention module. The fusion strategy based on the attention mechanism will give more weight to important features, and this will be more reasonable than the fusion strategy with directly specified weights. Specifically, our fusion strategy is based on unity fusion attention, and we exploit both channel and spatial attention based on this efficient architecture. The demonstration of the UFA module shows in Fig. 2.

Given an intermediate feature map $F_{k} \in \mathbb{R}^{C \times H \times W} ,k \in 1, \cdots, K$, where $K$=2 is set as the number of input images. Then the channel attention maps  $M_{k}^{c} \in \mathbb{R}^{C \times 1 \times 1}$  and spatial attention map $M_{k}^{s} \in \mathbb{R}^{1 \times H \times W}$ can be obtained through different operations. The whole fusion process can be formulated as below:
\begin{equation}
	F_{k}^{\prime}=M_{k}^{c}\left(F_{k}\right) \otimes F_{k},
\end{equation}
\begin{equation}
	F_{k}^{\prime \prime}=M_{k}^{s}\left(F_{k}^{\prime}\right) \otimes F_{k}^{\prime},
\end{equation}
where the $\otimes$ denotes element-wise multiplication.  $F_{k}^{\prime \prime}$  is the final output of the UFA module. Besides, it should be noted that the final outputs of the UFA module are concatenated at the trail of module. Then, the concatenated features are fed into the feature fusion block(FFB) for channel compression and feature fusion. Finally, the output of the FFB is input to the reconstruction module to generate the fusion image.
\subsubsection{Channel attention block}
The channel attention provides more flexibility to handle different types of information, focusing more attention on the important channel information. The channel attention maps are produced by calculating the channel relationship between different image features. In order to effectively compute the channel attention, the average-pooling operation is used to aggregating spatial information. As we know, the average-pooling operation can compute the spatial statistics. We empirically confirm that the average-pooling operation greatly improves the representation power of neural networks. The computation process of features is formulated as below.
\begin{equation}
	A_{k}=A P\left(F_{k}\right),
\end{equation}
\begin{equation}
	M_{k}^{c}=\frac{e^{A_{k}}}{\sum_{i=1}^{K} e^{A_{i}}},
\end{equation}
where the $AP$($\cdot$) denotes average-pooling operation, and $A_{k}$ denotes the average-pooled features. Then the features are forwarded to a Softmax activation function to generate the channel attention maps $M_{k}^{c} \in \mathbb{R}^{C \times 1 \times 1}$.
\begin{figure*}
	\centering
	\includegraphics[width=0.90\textwidth]{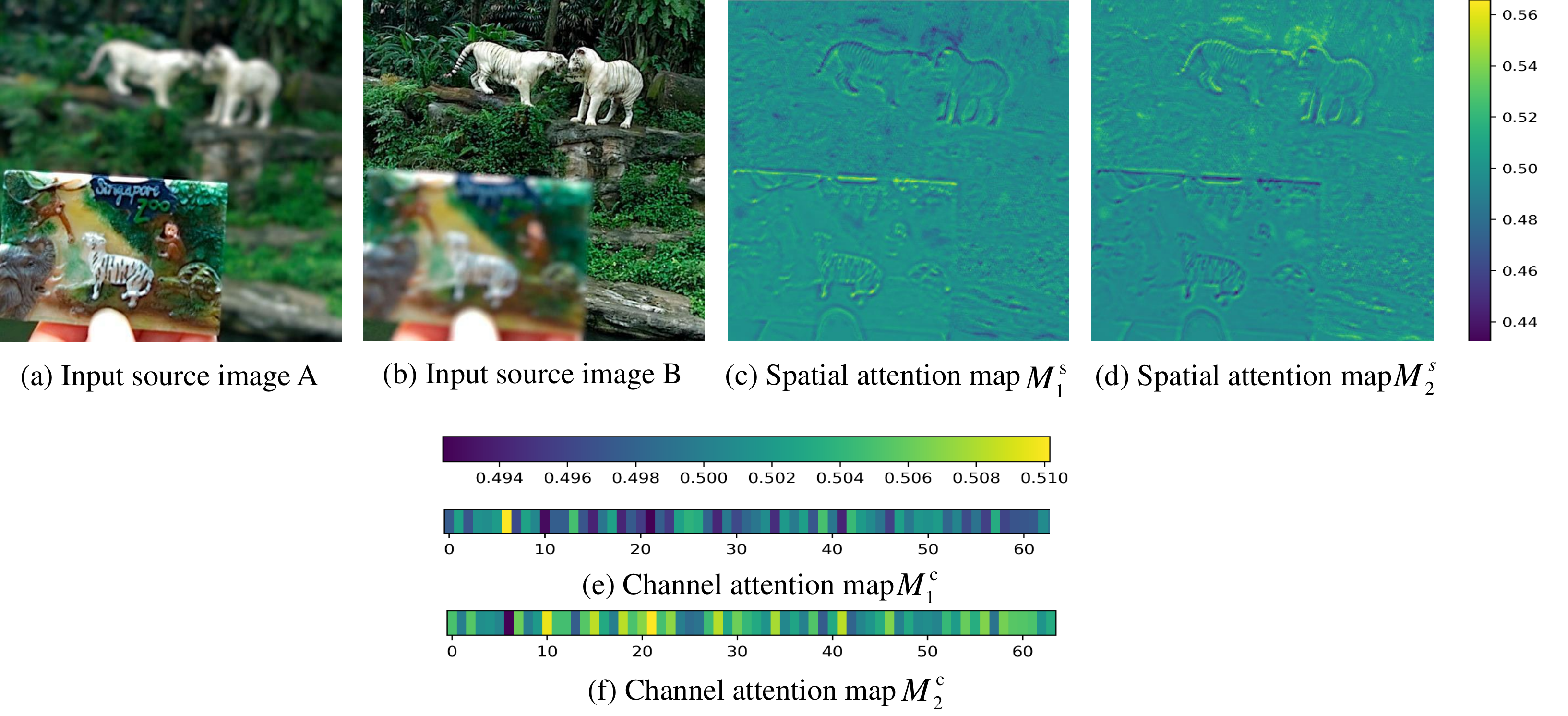}
	\vspace{-0.2cm}
	\caption{The Visualized attention feature maps. }
	\vspace{-0.5 cm}
\end{figure*}
\subsubsection{Spatial attention block}
 Spatial attention is different from channel attention, which concentrates on the informative region from the spatial dimension. The spatial attention maps are obtained from computing the spatial relationship of features. We apply the average-pooling operation and max-pooling operation in given image feature maps, and then the average-pooled features $F_{k}^{(s, a v g)} \in \mathbb{R}^{1 \times H \times W}$ and the max-pooled features $F_{k}^{(s, \max )} \in \mathbb{R}^{1 \times H \times W}$ are concatenated and the concatenated features are fed into a standard convolution layer to generate the spatial attention map $\mathit{AM}_{k}$. Afterward, the spatial attention map $\mathit{AM}_{k}$ are fed into the Softmax function to generate the final attention map $M_{k}^{s}$. 
\begin{equation}
	\begin{aligned}
		A M_{k} &=f^{7 \times 7}\left(\left[A v g P o o l\left(F_{k}^{\prime}\right); {MaxPool}\left(F_{k}^{\prime}\right)\right]\right),\\
		&=f^{7 \times 7}\left(\left[F_{k}^{(s,vy )} ; F_{k}^{(s,max )}\right]\right),
	\end{aligned}
\end{equation}
\begin{equation}
	M_{k}^{s}=\frac{e^{A M_{k}}}{\sum_{i=1}^{K} e^{A M_{i}}},
\end{equation}
where $f^{7\times7}$ denotes the standard convolution layer with the sizes of 7$\times$7. $AvgPool$ ($\cdot$) and $MaxPool$ ($\cdot$) represent the average-pooling operation and max-pooling operation, respectively.

To intuitively verify the effectiveness of the proposed unity attention mechanism, we visualize the channel-wise and spatial-wise feature weight maps of the output of the attention module. It can be clearly observed that the feature maps of different dimensions are adaptively learned with different weights. The proposed attention module treats different features unequally in different dimensions, while providing additional flexibility for the processing of different types of information. As shown in Fig. 3 (c) and (d), the spatial feature map $M_{1}^{\mathrm{s}}$ and $M_{2}^{\mathrm{s}}$ have greater weights at the boundaries as well as the salient target regions and edges within the image. This shows that the proposed attention mechanism makes the fusion module pay more attention to salient targets and edge regions from the spatial dimension during the fusion process. Fig. 3 (e) and (f) show the visualized channel weights maps, the output of the attention module in the channel dimension also clearly shows that different features can learn completely different weights adaptively. In conclusion, the proposed attention mechanism allows the fusion module to focus more on the salient target and edge regions during the fusion process.
\begin{figure*}
	\centering
	\includegraphics[width=0.90\textwidth]{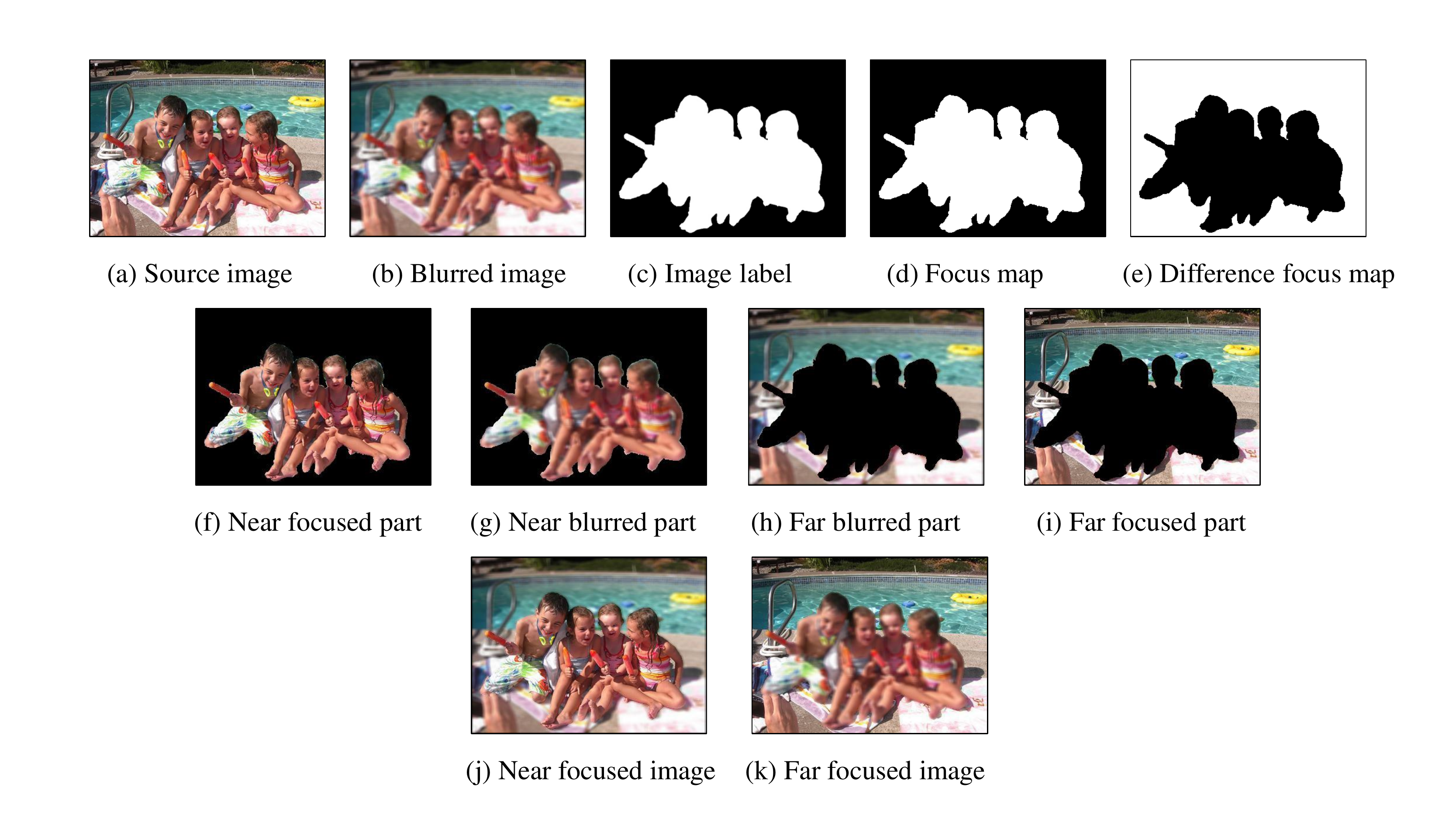}
	\vspace{-0.2cm}
	\caption{The demonstration of generating an multi-focus image dataset.}
	\vspace{-0.5 cm}
\end{figure*}
\subsection{Training dataset}
As is well-known, the models based on CNN are data-driven. Therefore, a well-generated large-scale image dataset plays an important role in the process of network training for deep learning-based models. In terms of training datasets, many works have tried different solutions. In \cite{liu2017multi}, Liu \emph{et al}. assumed that two states exist per pair of multi-focus image patches: the first state is that if one patch is focused and another patch is blurred, the second state is opposite to the first state. Based on this assumption, they created a huge image dataset which randomly cropped from the ImageNet dataset. The generated dataset contains approximately 2,000,000 pairs of image patches with sizes 16$\times$16. Besides, all the blurred patches in the generated dataset are obtained by blurring the focused patches with different levels of Gaussian filter. Similar to CNN, Tang \emph{et al}. formulated multi-focus image fusion as a classification task in p-CNN \cite{tang2018pixel}, they broadly divided the image patches into three categories: focused, blurred, or unknown. Therefore, they generated a large-scale image dataset which was rendered with 12 handcrafted blurring masks. This generated dataset has a total of 1450000 image patches with sizes 32$\times$32, specifically, it contains 650000 focused patches, 700000 blurred patches and 100000 unknown types of patches. Lacking large-scale dataset is not unique in other image fusion fields as well. For instance, the training dataset in DeepFuse \cite{prabhakar2017deepfuse} is obtained by randomly cropped image patches from another small scale multi-exposure dataset. What’s more, the dataset applied in DenseFuse \cite{li2018densefuse} is the MS-COCO dataset \cite{lin2014microsoft} which is generated for object detection. But the dataset is not specific and deliberate to train the infrared and visible image fusion model, which will absolutely limit the performance of the network.\\
\indent As we have learned in \cite{liu2017multi}, the dataset for training fusion model proposed in CNN only contains one type of image patch pairs (i.e., if one patch is blurred and another patch in the same image patch pair is focused). In addition, the same question exists in p-CNN, the training image patches have been classified into focused, unknown and defocused these three types to train the fusion model. We knew that these two datasets applied on CNN and p-CNN are essentially used to train classification tasks. Therefore, these two datasets are not appropriate for training the end-to-end image fusion network with supervised learning strategy. Compared to the two datasets mentioned above, the datasets in DeepFuse and DenseFuse are suitable for training an end-to-end fusion framework. However, the dataset applied in DeepFuse is obtained by randomly cropped on a small-scale multi-exposure image set, and the generated dataset lacks ground-truth fusion images. The DeepFuse fusion model has to be trained with the unsupervised learning strategy. Similar to DeepFuse, the training dataset in DenseFuse also lacks the ground-truth fusion image. DenseFuse trained on the MS-COCO dataset with the unsupervised SSIM \cite{wang2004image} loss function. It is no doubt that no ground-truth fusion image will absolutely restrict the performance of the trained fusion models. Besides, most datasets have been cropped with the small sizes 16$\times$16, 32$\times$32 and 64$\times$64 in the pretreatment processing according to previous works. Due to the small size of training image patches, it will lose some texture and contextual information in the training process of fusion model. This is not suitable for image fusion, since image fusion tasks need discriminative image features. \\
\indent As shown in the previous Refs. \cite{liu2017multi,prabhakar2017deepfuse,li2018densefuse}, a befitting dataset is important for image fusion models. To our minds, a qualified multi-focus image dataset should have at least two basic points: one point is that the ground-truth fusion images of the multi-focus images should be obtained from the generated dataset. Another point is that the number and the size of composite multi-focus images cannot be small and the composite should simulate real multi-focus images as much as possible. Inspired by previous works\cite{liu2017multi,prabhakar2017deepfuse,li2018densefuse}, we proposed a new way to generate a multi-focus image dataset based on the DUTS dataset reasonably and intuitively. DUTS \cite{wang2017learning} dataset is proposed for salient object detection at first, which consists of 15572 labeled images. Specifically, our multi-focus image dataset is generated from the DUTS dataset as the following steps:
\begin{figure*}
	\centering
	\includegraphics[width=0.90\textwidth]{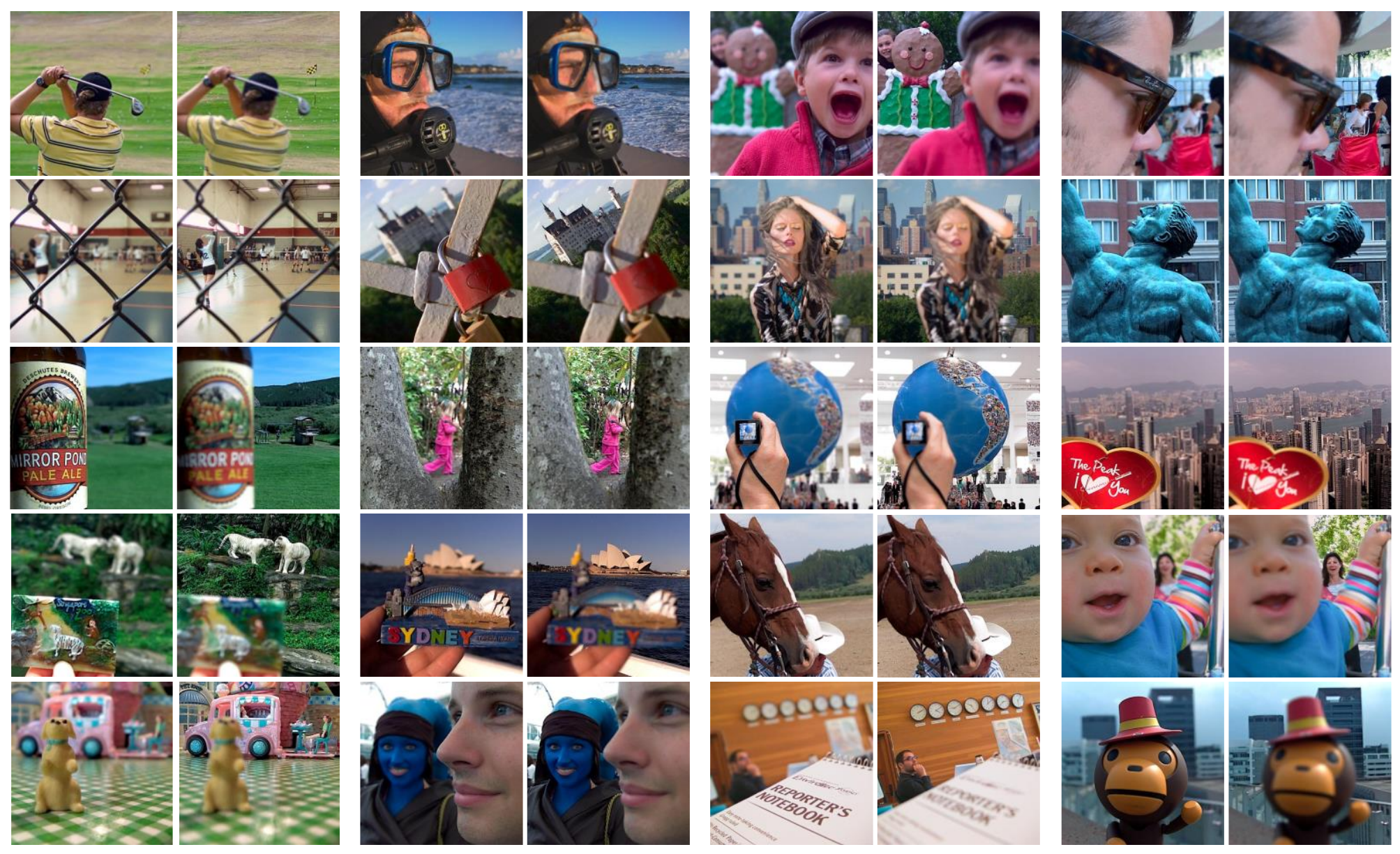}
	\vspace{-0.2cm}
	\caption{The multi-focus image dataset.}
	\vspace{-0.5cm}
\end{figure*}

(i) The source image $I_{s}$ from the DUTS dataset is blurred by the mean filter, and then the complete blurry image $I_{b}$ is obtained. Besides, when we use the mean filter with different sizes, we can obtain the images with varying degrees of blur.

(ii) Finding the focus map is an important procedure for the dataset generation method. The focus map is obtained from the image label, but the image label is not a binary map. Therefore, we choose a threshold function to normalize the image label into a binary map (i.e., focus map $I_{f}$).

(iii) Then a pair of multi-focus images could be generated according to source image $I_{s}$, blurry image $I_{b}$ and focus map $I_{f}$, as formulated below
\begin{equation}
	\left\{\begin{array}{l}
		I_{near}=I_{s} \odot I_{f}+I_{b} \odot\left(1-I_{f}\right), \\
		I_{far}=I_{b} \odot I_{f}+I_{s} \odot\left(1-I_{f}\right),
	\end{array}\right.
\end{equation}
where $I_{near}$ denotes the near focused image, and $I_{far}$ denotes the far focused image, they are a pair of multi-focus images, $\odot$ means dot product operation. 1 stands for an all one matrix and it has the same size with $I_{s}$ and $I_{f}$.

The multi-focus image dataset can be obtained according to these generation procedures. What’s more, as illustrated in step (i), we adopted the mean filter with different sizes to blur the image to make the generated dataset more natural. We set the size of the mean filter to four different intervals, which are [2, 3], [4, 5], [5, 7] and [8, 10], respectively. So, the generated dataset consists of four groups of images with different levels of blur. In addition, the number of our multi-focus image pairs is theoretically infinite according to this generation method. Overall, the dataset based on our generation method is more diverse and closer to reality than the previous works. Our dataset generation method shows in Fig. 4, which visualizes the state of the image after each generation procedure. Fig. 4 (a) is the source image and it is also taken as the ground-truth fusion image in the generated dataset. The blurred image shown in Fig. 4 (b) is generated through step (i), and the image label shown in Fig. 4 (c) is obtained from the original DUTS dataset. Then the focus map shown in Fig. 4 (d) is generated by step (ii) and the difference focus map, as expressed in Fig. 4 (e), is obtained by all one matrix subtracting the focus map. Finally, according to the step (iii), we would further obtain the near-focused multi-focus image and far focused multi-focus image.

In general, our generated multi-focus image dataset has three advantages compared to the previous dataset, for instance: (I) It is easy to implement when finding a salient objects detection dataset with image labels. (II) It can generate a sufficient number of multi-focus image pairs according to the training needs. (III) It owns ground-truth fusion images for each generated multi-focus image pairs. Based on these characteristics, we can train an end-to-end image fusion network based on our generated dataset.

\section{Experimental results and analysis}
In this section, we have conducted extensive experiments on publicly available datasets to validate the performance of the proposed image fusion model from a qualitative and quantitative perspective. Firstly, some experimental details are introduced. Then, the ablation study is conducted to discuss the setting of parameter $\lambda$ in the loss function and validate the effectiveness of the proposed unity fusion attention. Afterward, the qualitative evaluation results and quantitative evaluation results are illustrated. Finally, we evaluate the performance of other image fusion tasks to further demonstrate the effectiveness of our proposed fusion model.

\subsection{Experimental settings}
First of all, we introduce the training details which are important for the training of fusion models. The proposed fusion model is trained by using the generated dataset, which consists of 84424 images. In each training batch, we randomly extract 8 images with the sizes of 256$\times$256. We implement our proposed fusion model with Pytorch framework and update it with Adam optimizer \cite{kingma2014adam}. The learning rate is initialized to 1e-4 for all layers and divided into ten for every 200 epochs. The proposed fusion framework is trained for a total of 300 epochs, and the total training time is 60h. The proposed model is trained and tested on an eight-core PC with an Intel 9700KF 3.5 GHz CPU (with 16 GB RAM) and a GTX 2070Super GPU (with 8GB memory).

To objectively verify the superiority of our proposed fusion method over other fusion methods, we have compared it with nineteen representative fusion algorithms. What’s more, the comparison algorithms we chose are diverse and not limited to those deep learning-based methods. Specifically, the comparison algorithms contain such as sparse representation (SR) \cite{yang2009multifocus}, curvelet transform (CVT) \cite{nencini2007remote}, discrete wavelet transform (DWT)\cite{li1995multisensor}, dual-tree complex wavelet transform (DTCWT), non-subsampled contourlet transform (NSCT) \cite{zhang2009multifocus} these transform domain-based fusion algorithms; dense SIFT (DSIFT) \cite{liu2015multi}, guiding filter fusion (GFF) \cite{li2013image}, matting fusion (IMF)\cite{chen2017robust}, multi-scale weighted gradient fusion (MWGF) \cite{zhou2014multi} and spatial frequency (SF) \cite{li2008multifocus} these spatial domain-based algorithms; convolutional neural network (CNN) \cite{liu2017multi}, DeepFuse \cite{prabhakar2017deepfuse}, SESF, DenseFuse \cite{li2018densefuse} (including elementwise-add and $L_{1}$ fusion strategy), DRPL\cite{9020016}, MFF\cite{ZHANG202140} and IFCNN \cite{zhang2020ifcnn} (including elementwise-maximum fusion strategy) these deep learning-based algorithms. Compared with these traditional comparison algorithms (including transform domain-based and spatial domain-based algorithms), the proposed algorithm has stronger feature representation capability and can effectively extract more informative image features. Moreover, compared with these deep learning-based comparison algorithms, the proposed algorithm is an end-to-end network structure and does not require the generation of an intermediate decision map to achieve image fusion. In addition, the proposed algorithm designs the novel attention mechanism-based fusion strategies, which can provide additional flexibility for fusing different types of image features.\\
\begin{figure*}
	\centering
	\includegraphics[width=0.9\textwidth]{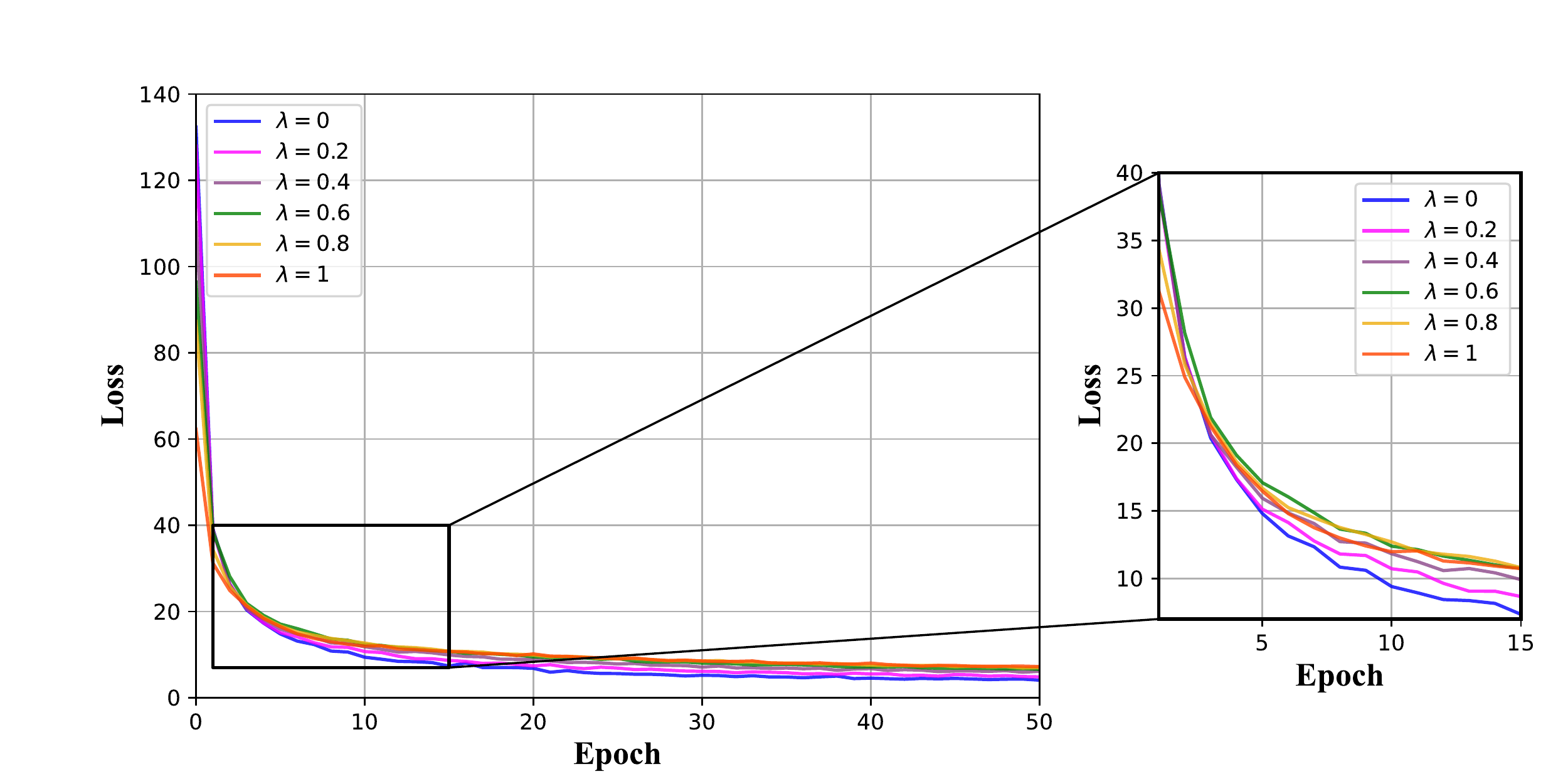}
	\vspace{-0.2cm}
	\caption{The line charts of Loss in training phase.}
	\vspace{-0.5 cm}
\end{figure*}
\indent What’s more, we have discriminated against the performance of different fusion models through qualitative and quantitative evaluation methods. From the perspective of qualitative evaluation methods, whether the visual effects of the fusion image differ from the source images is the main discriminating method. Specifically, the fused image should contain as many distinct and sharp features from source images in multi-focus image fusion, and does not exist additional artifacts.   Since only evaluating the performance of fusion methods from the qualitative perspective is not comprehensive. Consequently, we further use seven metrics to evaluate the performance of the fusion algorithms on the multi-focus image fusion. The seven metrics are respectively the visual information fidelity (VIFF) \cite{han2013new}, standard deviation (STD) \cite{liu2005image}, average gradient (AVG) \cite{song2007fusion}, gradient-based fusion performance ($Q^{abf}$) \cite{emmerich2007gradient}, shannon entropy (SEN) \cite{wax1977improved}, $Q^{c}$ \cite{chen2009new} and $Q^{m}$ \cite{wang2008novel}. \\
\indent VIFF, which is an evaluation metric based on human visual perception, measures the visual information fidelity between the fusion image and source images. STD and AVG, which are two image features-based evaluation metrics, measure the textural information of the fusion image from a statistical point of view. $Q^{abf}$ is an evaluation metric based on source images and the fusion image, and $Q^{abf}$ evaluates the performance of fusion methods based on gradient. SEN is an evaluation metric based on information theory. $Q^{m}$ retrieves the edge information from the high and band-pass components of the decomposition by using the two-level Haar wavelet. $Q^{c}$ employs the major features in human visual system model to evaluate the perceptual quality of image fusion. There is no doubt that these seven metrics cover different types of fusion evaluation metrics. So, it is enough to evaluate the fusion results of different fusion models through these seven metrics. These seven metrics can objectively reflect the abilities of fusion methods on merging the clear image features and on integrating the texture information. It is certainly appropriate to choose these seven metrics to evaluate the performance of different fusion algorithms. In addition, it should be noted that qualitative evaluation and quantitative evaluation perform on the ‘Lytro’ multi-focus image dataset \cite{nejati2015multi}, as shown in Fig. 5, the dataset contains 20 pairs of multi-focus images.

\begin{figure*}
	\centering
	\includegraphics[width=0.90\textwidth]{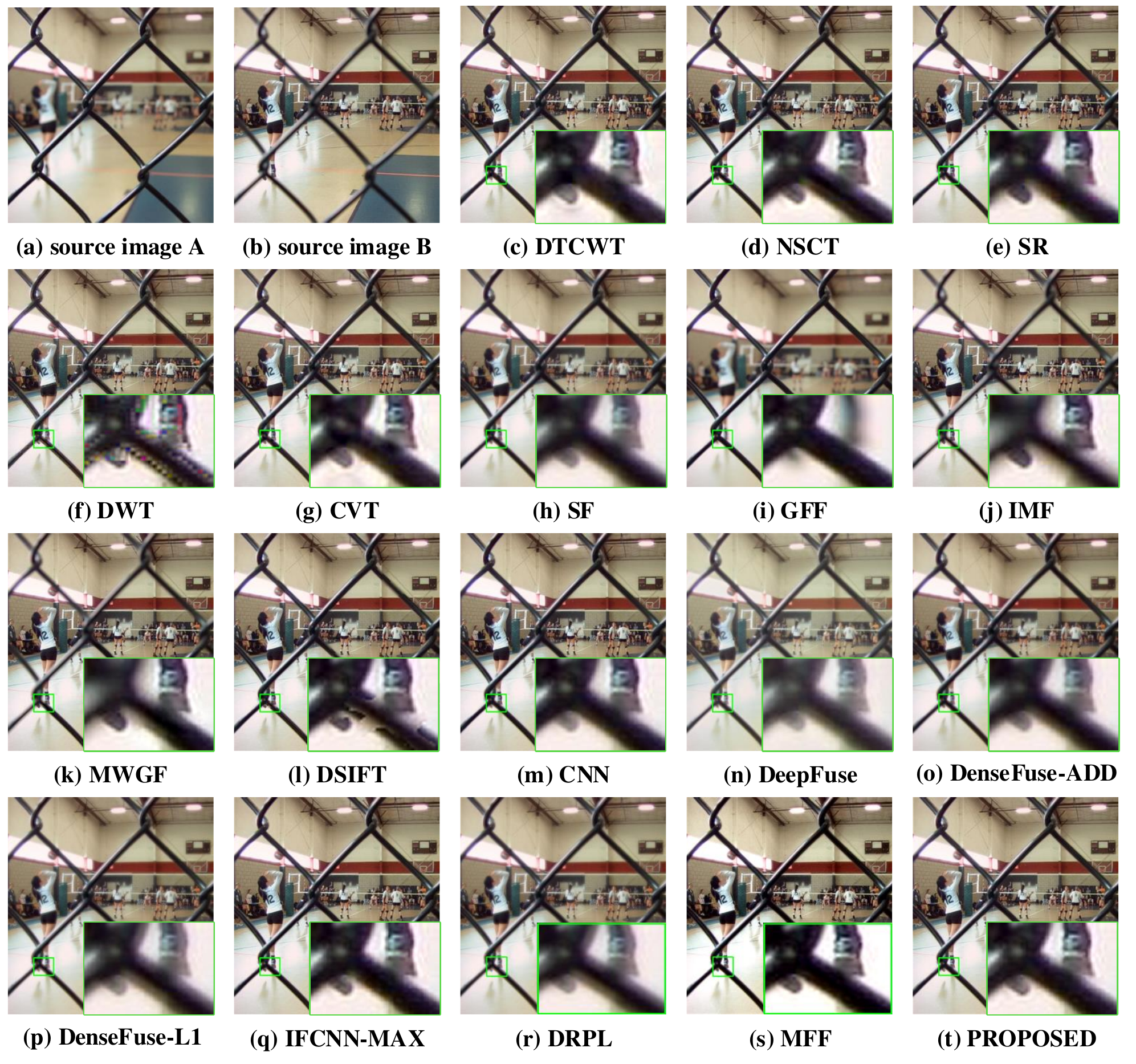}
	\vspace{-0.2cm}
	\caption{The visualization of “Volleyball court” fused results.}
	\vspace{-0.5cm}
\end{figure*} 
\begin{figure*}
	\centering
	\includegraphics[width=0.9\textwidth]{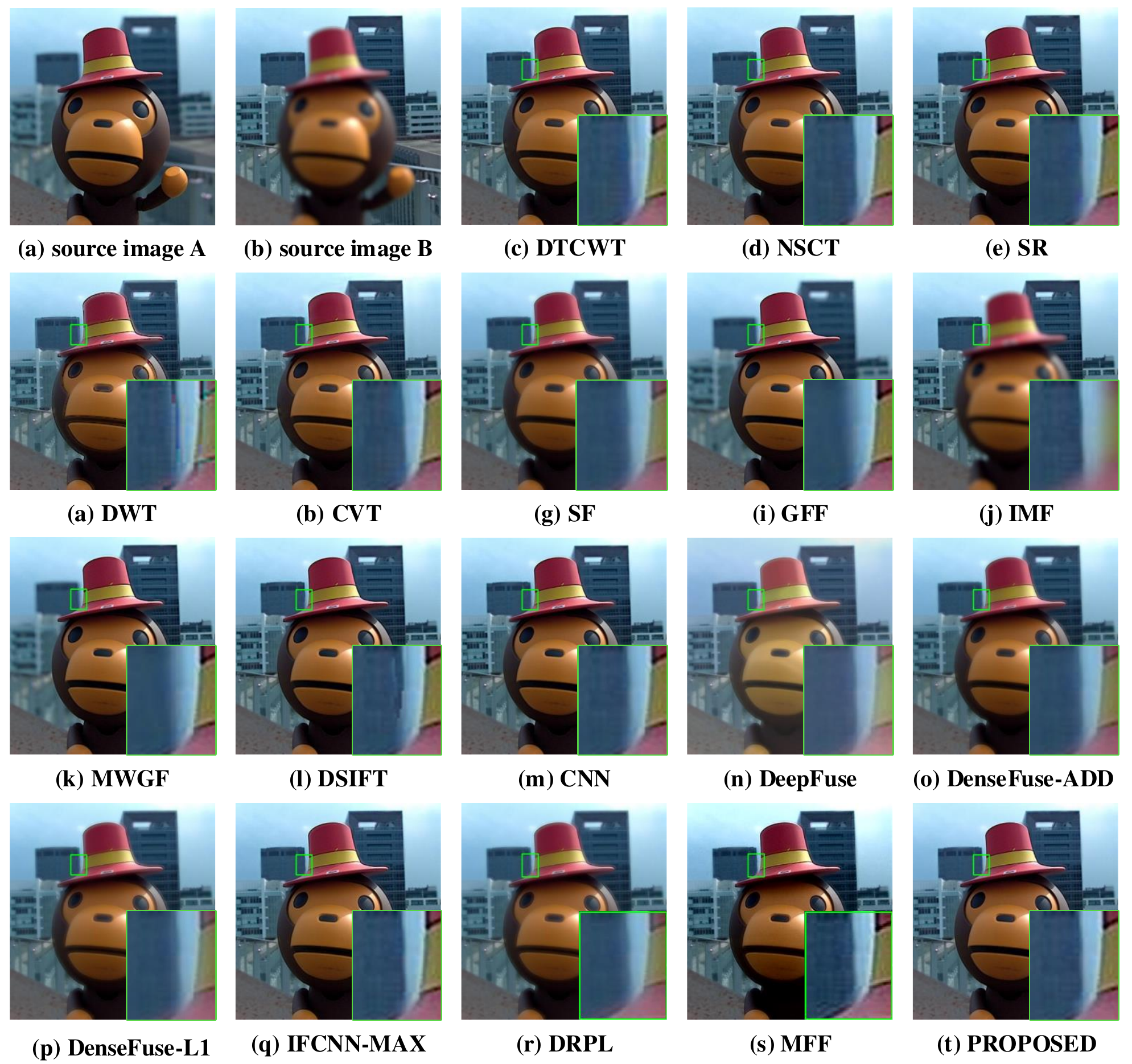}
	\vspace{-0.2cm}
	\caption{The visualization of “Toy monkey” fused results.}
	\vspace{-0.5cm}
\end{figure*}
\begin{figure*}
	\centering
	\includegraphics[width=0.9\textwidth]{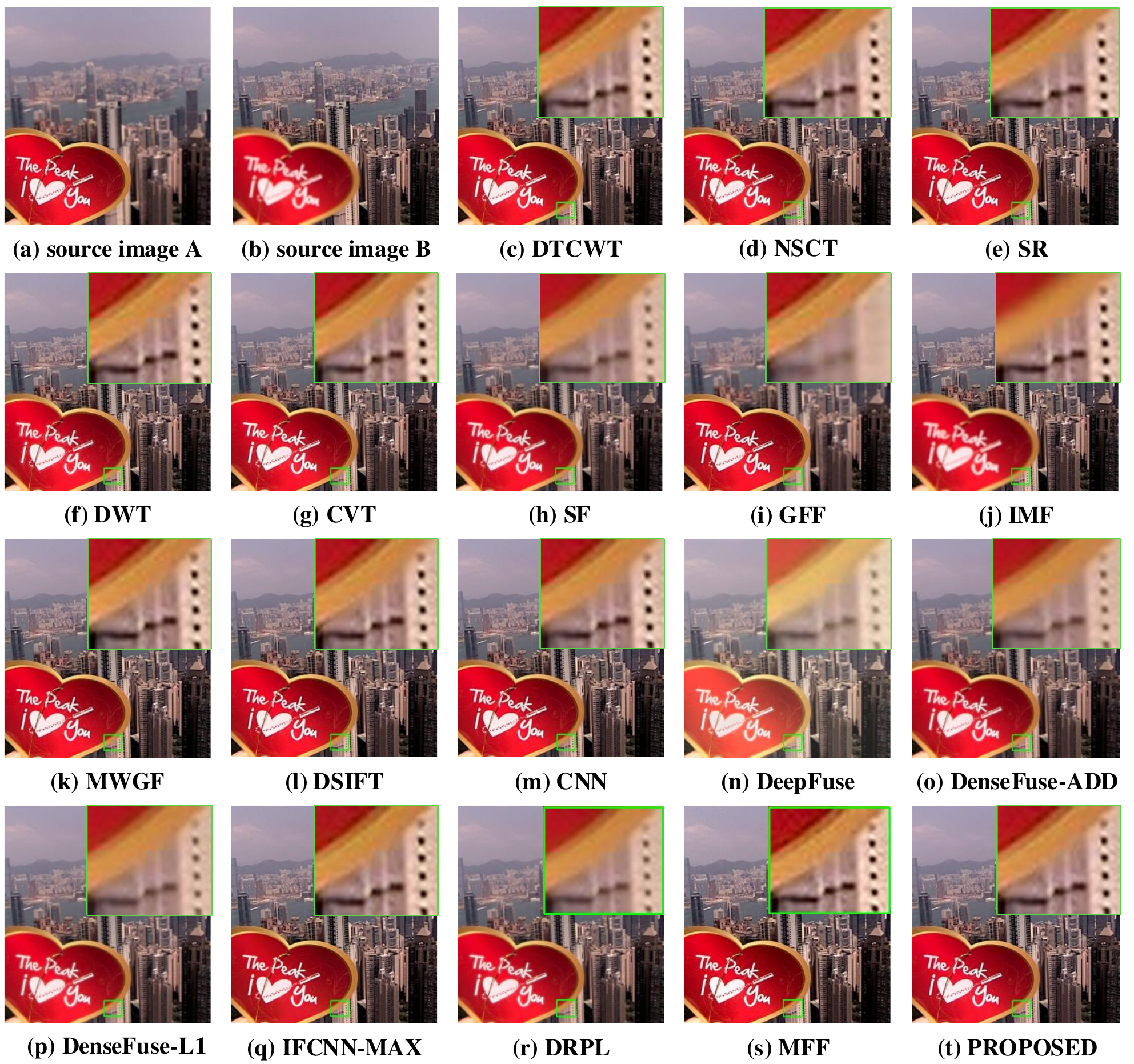}
	\vspace{-0.2cm}
	\caption{The visualization of “Heart-shaped label” fused results.}
	\vspace{0 cm}
\end{figure*}

\subsection{Ablation study}
\subsubsection{Parameter($\lambda$) in Loss Function}
As stated in  Section III-A, the value of the parameter $\lambda$ in the loss function \emph{L} is set to 0.2 with reference to the priori empirical value\cite{7797130}. We need to experimentally verify whether the parameter $\lambda$ is appropriate when it takes the value of 0.2. Therefore, we trained the network by taking different values of the parameter $\lambda$ and visualized the loss function curves of the network with different parameters $\lambda$. During the experiments, the parameters $\lambda$ were set to 0, 0.2, 0.4, 0.6, 0.8 and 1. The visualized loss function curves are shown in Fig. 6. In Fig. 6, we can see that the convergence of the fusion network slows down as the parameter $\lambda$ increases. When $\lambda$ = 0 or $\lambda$ = 0.2, the convergence speed of the fusion network is the fastest and the two curves are almost indistinguishable. Furthermore, to verify which parameter $\lambda$ makes the fusion network achieve the best results. We selected four metrics to quantify the performance of the network with different values of $\lambda$ on the test image set after 50 epochs of training. These quantified values are shown in Table III, with the best value in green and the second-best value in blue. As shown in Table III, we can clearly see that the fusion network achieves better fusion results when $\lambda$ is taken to be 0.2. This also shows that it is appropriate to take a value of 0.2 for the parameter $\lambda$ in our experiments with reference to the priori empirical value.
\begin{table}[t]
	\centering
	\caption{THE METRICS VALUES WITH DIFFERENT $\lambda$}
	\begin{tabular}{lcccc}
		\hline
		\multicolumn{1}{r}{} & AVG    & STD     & SEN    & $Q^{a b f}$   \\ \hline
		$\lambda=0$                     & 8.1440 & 62.2940 & 7.5523 & \color{green}0.7312 \\
		$\lambda=0.2$                    & \color{blue}8.2149 & \color{green}62.6244 & \color{green}7.5689 & \color{blue}0.7300 \\
		$\lambda=0.4$                    & 8.1690 &\color{blue} 62.5130 & \color{blue}7.5608 & 0.7214 \\
		$\lambda=0.6$                    & 8.1764 & 62.5121 & 7.5584 & 0.7278 \\
		$\lambda=0.8$                    & \color{green}8.2644 & 62.2040 & 7.5427 & 0.7228 \\
		$\lambda=1$                      & 8.0715 & 61.7889 & 7.5570  & 0.7240 \\ \hline
	\end{tabular}
	\label{tbl:table-example}
\end{table}
\subsubsection{The influence of the proposed attention model}
We evaluated our proposed fusion model with ablation experiments to verify the contribution of the designed unity fusion attention module. We picked four different modes to train the image fusion model with 50 epochs. There are ‘NO-UFA’, ‘NO-SA’, ‘NO-CA’ and UFA, respectively. ‘NO-UFA’ denotes that the fusion model does not contain the unity attention module. ‘NO-SA’ denotes the UFA module in the fusion model has removed the spatial attention module. ‘NO-CA’ denotes that the UFA module in the fusion model has been removed from the channel attention module. ‘UFA’ denotes our proposed fusion model which contains a complete UFA module. For a more intuitive comparison, we quantify the performance of these four trained modes on the ‘Lytro’ multi-focus image dataset through the metrics of AVG, SEN and STD. Table IV shows the quantitative results of these four fusion modes. It can be seen from Table IV that the fusion mode obtains the relatively low value on three evaluation metrics when the UFA module is removed. Besides, we can clearly observe that the fusion modes with CA or SA could perform better than the fusion mode without CA or SA. Of course, there is no doubt that the fusion mode with UFA module achieves the best performance on three evaluation metrics. These comparisons firmly demonstrate the effectiveness of UFA and indicate the fusion mode with UFA is more effective than the fusion mode with CA or SA. In summary, the proposed UFA module is applicable to achieve effective and powerful fusion network for image fusion.
\begin{table}[t]
	\centering
	\caption{The quantitative results of ablation study}
	\begin{tabular}{cccccc}
		\hline
		\vspace{-0.1cm}
		& AVG & SEN & STD \\
		\midrule
		NO-UFA & 8.1859 & 7.5629 & 61.6871\\
		NO-SA & 8.1956 & 7.5660 & 61.8769\\
		NO-CA & 8.1964 & 7.5668 & 62.2399\\
		UFA &\color{green} 8.2149 &\color{green} 7.5689 &\color{green} 62.6244\\
		\hline
	\end{tabular}
	\label{tbl:table-example}
\end{table}
\begin{table*}[h]
	\centering
	\caption{The quantitative results of comparison algorithms}
	\setlength{\tabcolsep}{2.5mm}{
		\begin{tabular}{ccccccccc}
			\hline
			Methods & AVG & STD & SEN & VIFF & $Q^{abf}$ & $Q^{c}$ &$Q^{m}$ & Time \\
			\midrule
			
			DWT(Li and Manjunath \emph{et al}. 1995) & 8.3285 & 61.6849 & 7.5676 & 0.8730 & 0.6839 & 0.7925 & 2.9613 &11.34\\
			CVT(Nencini \emph{et al}. 1995)	 & 8.2652 & 62.2107 & 7.5701 & 0.9359 & 0.7113 & 0.7993 & 2.9662 & 12.98\\
			DTCWT(Lewis \emph{et al}. 2007) & 8.3016 & 62.2172 & 7.5682 & 0.9368 & 0.7245 & 0.8077 & 2.9693 & 11.68\\
			NSCT(Zhang and Li \emph{et al}. 2009) & 8.2761 & 62.3538 & 7.5661 & 0.9524 & 0.7250 & 0.8160 & 2.9682 & 17.94\\
			SR(Yang an Li \emph{et al}.2010) & 8.2142 & 62.2011 & 7.5656 & 0.9510 & 0.7299 & 0.8132 & 2.9704 & 155.65\\
			SF(Li and Kork \emph{et al}. 2011) & 5.2123 & 59.8217 & 7.5421 & 0.7085 & 0.5854 & 0.7905 & 2.8363 & 1.78\\
			GFF(Li and Kang \emph{et al}. 2014) & 5.4857 & 60.5329 & 7.5468 & 0.7489 & 0.5651 & 0.7921 & 2.8423 & 28.32\\
			IMF(Chen and Guan \emph{et al}. 2015) & 6.3718 & 60.8008 & 7.5462 & 0.7644 & 0.6189 & 0.7135 & 2.8756 & 30.55\\
			MWGF(Zhou \emph{et al}. 2014) & 8.1591 & 62.2509 & 7.5700 & 0.9598 & 0.7322 & 0.8119 & 2.9666 & 23.16\\
			DSIFT(Liu \emph{et al}. 2015) & 8.3142 & 62.3776 & 7.5675 & 0.9657 &\color{green} 0.7555 & 0.8139 & 2.9726 & 41.95\\
			CNN(Liu \emph{et al}. 2017) & 8.2537 & 62.3058 & 7.5694 & 0.9580 & 0.7394 &\color{blue} 0.8177 & 2.9707 & 96.73\\
			DeepFuse(Li \emph{et al}. 2017) & 4.8398 & 54.9187 & 7.3964 & 0.5718 & 0.4969 & 0.6936 & 2.8116 & 0.62\\
			DenseFuse-ADD(Li \emph{et al}. 2019) & 5.1821 & 59.6207 & 7.5391 & 0.7030 & 0.5834 & 0.7894 & 2.8350 & 3.31\\
			DenseFuse-L1(Li \emph{et al}. 2019) & 5.3261 & 59.4228 & 7.5385 & 0.6927 & 0.6030 & 0.7923 & 2.8379 & 19.40 \\
			IFCNN-MAX(Zhang \emph{et al}. 2020) & 8.0665 & 62.0851 & 7.5618 & 0.9562 & 0.7113 & 0.8176 & 2.9529 & null\\
			SESF(Ma \emph{et al}. 2020) &\color{blue} 8.3345 & 62.4297 & 7.5686 &\color{blue} 0.9676 & 0.7375 & 0.8160 &\color{blue} 2.9731 & 2.53\\
			DRPL(Li \emph{et al}. 2020) & 8.3086 & 62.3017 & \color{blue}7.5751 & 0.9643 & 0.7342 & 0.8102 & 2.9666 & \color{blue}0.33\\
			MFF(Zhang \emph{et al}. 2021) & 8.2952 & \color{blue}62.6967 & 7.4035 & 0.9673 & 0.7364 & 0.7444 & 2.9163 & 1.14\\
			Proposed & \color{green} 8.3401 &\color{green} 62.7020 &\color{green} 7.5791 &\color{green} 0.9679 &\color{blue} 0.7417 &\color{green} 0.8187 & \color{green} 2.9742 &\color{green}0.26 \\
			\hline
		\end{tabular}
		\label{tbl:table-example}}
\end{table*}

\subsection{Multi-focus image fusion}
Since our proposed model has been trained on the multi-focus image dataset we created, we firstly want to authenticate the performance of the trained model on fusing multi-focus images, and then compare it with other advanced multi-focus image fusion algorithms to verify the superiority of our proposed fusion model. We show three comparison examples in Fig. 7, Fig. 8 and Fig. 9.\\
\indent Fig. 7(a)-(t) show the fusion results of source images Fig. 7(a) and (b), the source images show the far focused volleyball court and the near focused chain-link fence, respectively. The fusion image is to integrate the clear visual information from source images into a completely clear image as much as possible. As shown in Fig. 7, not all the fusion algorithms have achieved desirable performance. It can be seen from Fig. 7(f) that DWT fails to fuse the near- focused chain-link fence, the partially enlarged screenshot clearly shows the fusion image losses some texture information about the chain-link fence. Fig. 7(d) shows that the fusion image of NSCT shows oblique artifacts in chain-link fence. Fig. 7(e), Fig. 7(h) and Fig. 7(r) show that the fusion image of SR, SF and DRPL are more blurry than other fusion images. As shown in Fig. 7(i)-(l), the fusion image of GFF, IMF, MWGF and DSIFT all have not well integrated the clear image features from the source images, blocks of blur and artifacts can be easily observed in closeups. It can be seen from Fig. 7(n)-(p) that DeepFuse and DenseFuse abnormally increase the luminance of the fusion image. As for Fig. 7(c), (g), (m), (q), (s) and (t), the fusion images of DTCWT, CVT, CNN, IFCNN, MFF and the proposed model all have well merged the clear visual information from source images, and showed pleasant visual effects that are different from other fusion algorithms.\\
\indent  In Fig. 8, we visualize the fused results of Fig. 8(a) and (b), which captured a toy monkey in front of two buildings. The ideal fusion image should combine the near-focused toy monkey and the far-focused two buildings. As shown in the magnified region which is placed in the lower-right corner in each image, Fig. 8(f), (j), (l) and (r) show that the fusion images of DWT, IMF, DSIFT and DRPL suffer from blurring effects around the toy monkey's hat. What's more, it can be seen from Fig. 8(d) that the boundary of the building is distorted. Fig. 8(h), (i) and (k) show that SF, GFF and MWGF are weak for integrating the clear image features from both source images. The building does not show clear boundaries. Besides, although the DeepFuse performs image fusion not bad, the fusion image as shown in Fig. 8(n) is brighter than other fusion images. Fig. 8(o) and (p) show that the fusion images of DenseFuse are low-contrast, which means that the DenseFuse lacks strong ability to fuse multi-focus images. Finally, as shown in Fig. 8(g), (m), (q), (s) and (t), CVT, CNN, IFCNN, MFF and our proposed model perform image fusion well, and achieves remarkable fusion effect.

Fig. 9(a)-(t) show the fusion results of images Fig. 9(a) and (b), which captured a heart-shaped label in front of buildings. As the closeups demonstrated, Fig. 9(h), (i), (n), (o) and (p) show that the fusion results of SF, GFF, DeepFuse and DenseFuse fail to fuse the clear image features at the border, the building looks blurry. It can be seen from Fig. 9(j) that the border of the heart-shaped label is almost blurred, which might be caused by excessive smoothing. What's more, it is easy to be disturbed by noise in the process of image fusion. We can observe that the fusion results of DWT, CVT, DTCWT, NSCT, DSIFT and DRPL exist point of blurring in the border of the building. In the end, our proposed model, CNN and IFCNN have finely integrated the clear image features into fusion images, generating comparable fusion images.

Besides the qualitative evaluation for each fusion result, we have also used the seven objective metrics to quantitatively evaluate the performance of the different fusion algorithms. Table V lists the quantitative results, each value in this table denotes the mean metric value on the full “Lytro” dataset which is fused by different fusion methods. Moreover, the best value is shown in green and the second-best value is shown in blue in the corresponding metric column of this table. We can see from the quantitative comparison in Table V that our proposed model and SESF clearly outperform the other 19 fusion methods on the value of the AVG, STD, SEN, VIFF, $Q^{c}$ and $Q^{m}$ fusing evaluation metrics, which implies our proposed model and SESF have obtained better image features, edge information and higher visual information fidelity. As for the $Q^{abf}$ metric, the DSIFT achieved the highest value and our proposed model ranks second. According to the all-inclusive evaluation on the public multi-focus image dataset, our proposed fusion model can sustain relatively higher visual information ﬁdelity, preserve more gradient and textural information, and produce visually-pleasing fusion image compared to other comparison algorithms. Thus, our proposed fusion model has demonstrated better performance on the multi-focus image dataset compared to other state-of-art algorithms.
In addition, we compare the time efficiency of each fusion framework by calculating the average time to process an image. Apparently, the average time for processing an image is the least for our proposed framework which takes only 0.26 seconds, while the fastest of all other comparison methods takes 0.33 seconds. This also indicates that the proposed framework has efficient processing efficiency while recovering high-quality fused images.
\begin{figure*}
	\centering
	\includegraphics[width=0.70\textwidth]{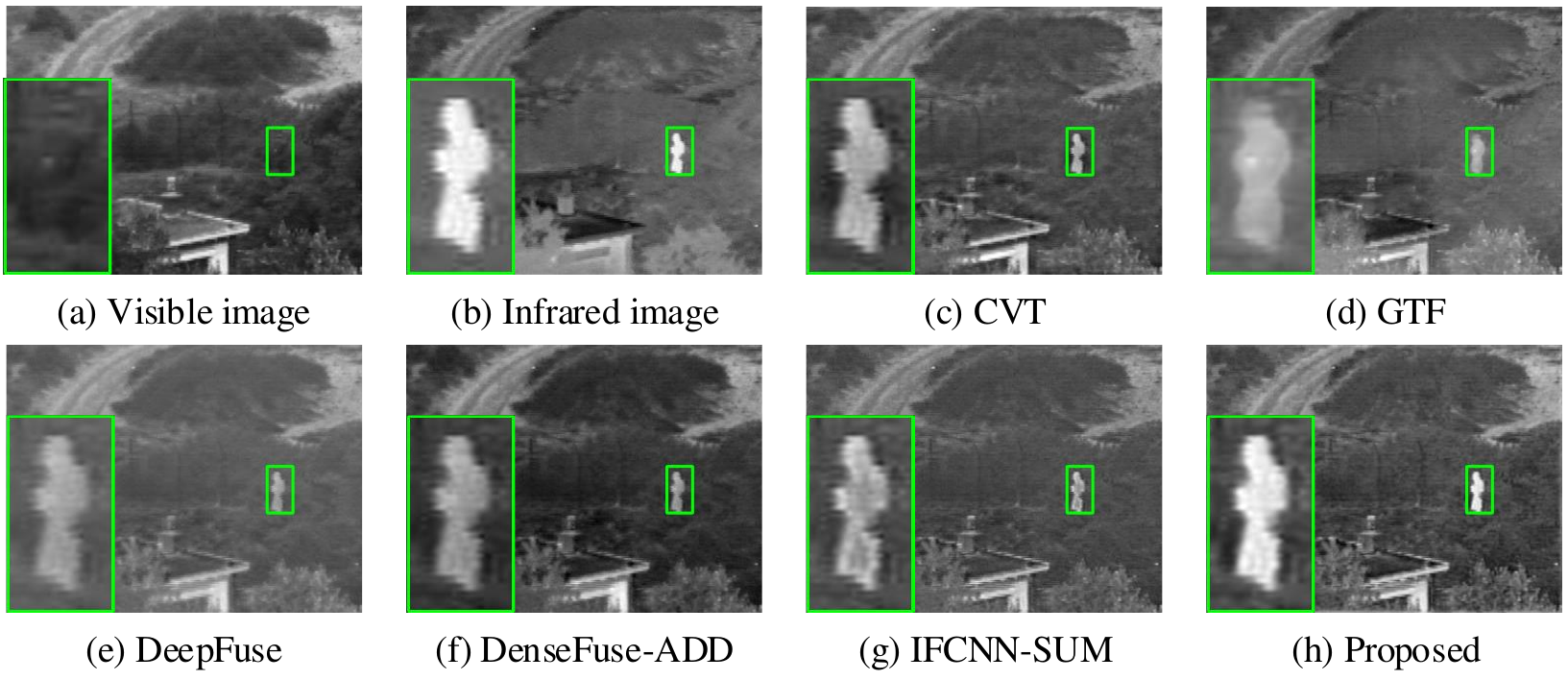}
	\vspace{-0.2cm}
	\caption{The comparison example of infrared and visible images.}
	\vspace{0 cm}
\end{figure*}
\begin{table*}[h]
	\centering
	\caption{the quantitative evaluation results on the visible and infrared dataset.}
	\begin{tabular}{ccccccccc}
		\toprule
		Methods & CVT & GTF & DenseFuse-ADD & DeepFuse & IFCNN-SUM & Proposed\\
		\midrule
		AVG & \color{blue} 4.558 & 3.014 & 3.810 & 2.921 & 4.525 &\color{green} 4.802\\
		STD & 25.551 & 24.157 &\color{green} 31.310 & 22.644 & 24.212 &\color{blue}  29.214\\
		VIFF & 0.311 & 0.257 & \color{green} 0.352 & 0.292 & 0.301 &\color{blue} 0.313\\
		$Q^{abf}$ &\color{blue}  0.509 & 0.412 & 0.456 & 0.392 &\color{green} 0.517 & 0.489\\
		\bottomrule
	\end{tabular}
	\label{tbl:table-example}	
\end{table*}
\subsection{Expanded experiments}
Our proposed fusion network also serves as a fusion model for other image fusion tasks. We evaluate the infrared and visible image fusion performance and medical image fusion performance to demonstrate the effectiveness of our proposed fusion model. Here we choose five representative algorithms to compare with our fusion model. Three of these five comparison algorithms are designed for infrared and visible image fusion and medical image fusion. They are GTF\cite{ma2016infrared}, DenseFuse-ADD \cite{li2018densefuse} and IFCNN-SUM \cite{zhang2020ifcnn}. As for CVT \cite{nencini2007remote} and DeepFuse \cite{prabhakar2017deepfuse}, they are two general image fusion methods which can be applied in different image fusion tasks. Besides, we choose four metrics from seven metrics which are mentioned above to quantify the performance of these algorithms on infrared and visible image fusion and medical fusion. They are AVG, STD, VIFF and $Q^{abf}$, respectively. To objectively verify the performance of these methods, we also evaluate the performance of these fusion tasks from qualitative and quantitative perspectives. 

\subsubsection{Infrared and visible image fusion}
Fig. 10 shows the comparison examples, and the source visible and infrared images are from the ‘TNO’ visible and infrared dataset \cite{zhou2016perceptual} which includes 14 image pairs. It is well known that a good fusion image should integrate as much as visible image features from visible image and the salient bright image features from the infrared image\cite{9127964}. It can be seen from Fig. 10 (c)-(g) that the fusion images of CVT, GTF, DeepFuse, DenseFuse-ADD and IFCNN-SUM fail to preserve the bright image features, the infrared imaging of the human figures which can be observed in closeups are dim compared to the source infrared image. However, our proposed fusion model has integrated the salient bright image features into the fusion image successfully. Fig. 10 (h) shows that the people figure in fusion image is brighter than other methods. Although our proposed fusion model achieves good performance compared to other algorithms, there are some shortcomings in the fusion image. Our proposed fusion model is not designed for infrared and visible image fusion, so the fusion model does not integrate the meaningful image feature from the source image completely. The trees in the fusion image are not as clear as in the visible image, and this problem also exists in other fusion images. Broadly speaking, our proposed fusion model achieves the best performance in comparison algorithms.\\
\indent Table VI shows the performance of the fusion algorithms on the ‘TNO’ infrared and visible image dataset. Each metric value listed in Table VI is the mean metric value on the full infrared and visible image dataset. In the evaluation of four metrics, our fusion model outperforms other methods. Our fusion model obtains the high value on AVG, STD and VIFF, these high metric values on these three metrics in consistence to our visual judgment. Moreover, the high values on metrics mean that our fusion model can integrate more rich information and texture information from source images into the fusion image. Absolutely, the thing worth commenting on is that the correlation between the visible image and infrared image is relatively low\cite{hou2020vif}, which differs from the multi-focus images. Therefore, a poorly fused image is more similar to the source image because it integrates only part of the image's features, while a well-fused image is more different from the source image because it integrates most of the source image's features. This is why IFCNN-SUM, which is poorly fused, obtains a higher value for the metric $Q^{abf}$ based on the evaluation of source images and the fusion image. 
\begin{figure*}
	\centering
	\includegraphics[width=0.70\textwidth]{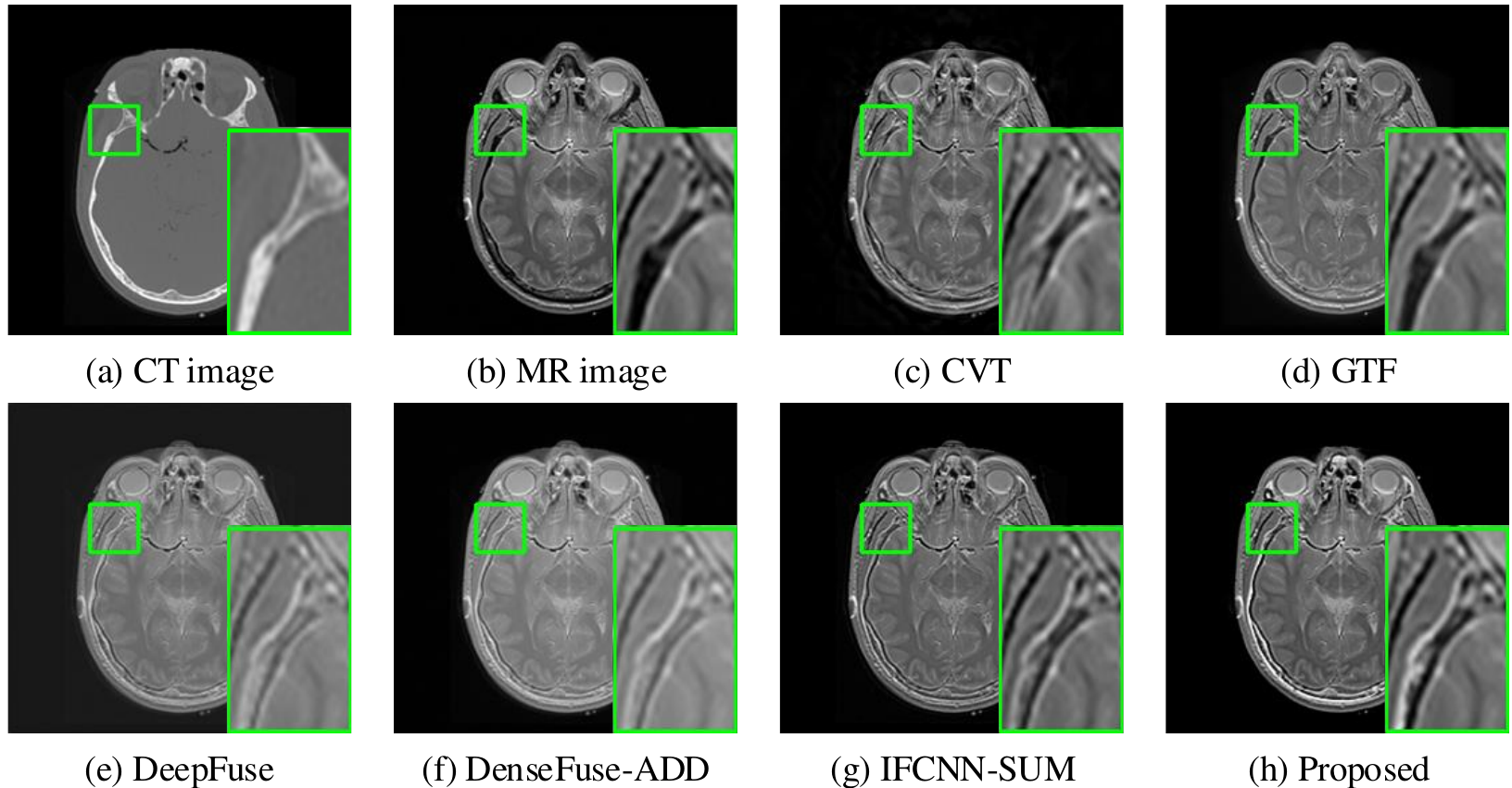}
	\vspace{-0.2cm}
	\caption{The comparison example of medical images.}
	\vspace{0 cm}
\end{figure*}
\begin{table*}[h]
	\centering
	\caption{the quantitative evaluation results on the medical image dataset.}
	\begin{tabular}{ccccccccc}
		\toprule
		Methods & CVT & GTF & DenseFuse-ADD & DeepFuse & IFCNN-SUM & Proposed\\
		\midrule
		AVG & 8.718 & 6.433 & 5.976 & 5.919 & \color{blue} 8.949 &\color{green} 9.003\\
		STD & \color{blue} 58.272 & 50.522 & 57.561 & 51.518 & 57.611 &\color{green} 67.270\\
		VIFF & 0.243 & 0.250 &\color{green} 0.287 & 0.257 & 0.257 &\color{blue}  0.264\\
		$Q^{abf}$ & \color{blue} 0.523 & 0.450 & 0.464 & 0.391 &\color{green} 0.543 & 0.479\\
		\bottomrule
	\end{tabular}
	\label{tbl:table-example}	
\end{table*}
\subsubsection{Medical image fusion}
In this subsection, we also evaluate the performance of our proposed model performs on the multi-modal medical image dataset \cite{liu2017medical}. The multi-modal medical image dataset contains eight pairs of images, namely eight CT images and eight MR images. Fig. 11 (a) and (b) show a pair of CT and MR slices which are obtained by scanning the human brain. As everyone knows\cite{singh2019multimodal,li2020laplacian}, an ideal medical fusion image should preserve as much as textural tissue features from the source MR image and bright skull features from the source CT image. As shown in Fig. 11 (c) and (d), the fusion image of CVT and GTF fails to integrate some portions of bright skull features from the source CT image. It can be seen from Fig. 11 (e) and (f) that DeepFuse and DenseFuse-ADD have successfully injected the salient image features from source CT and MR images into the fusion image, but the luminance of the fusion images has changed a lot compared to the source images. Fig. 11 (g) and (h) show that IFCNN-SUM and our proposed model have successfully integrated most of the textural tissue and bright skull features from source images. Besides, the fusion image generated by our proposed fusion model has a better visual quality than IFCNN-SUM, which can be easily observed in closeups.
 
Table VII shows the evaluation results of the fusion algorithms on the medical image dataset, and each value listed in Table VII is the mean metric value on the total eight pairs of medical images. As shown in Table VII, our proposed fusion model is superior to other fusion algorithms, such as the quantitative evaluation results of AVG and STD rank in the first place, the quantitative evaluation result of VIFF ranks in second place, which indicates our proposed fusion model could inject more textural and bright image features from source CT and MR images into the fusion image. Similar to the infrared and visible images, the medical images with different modalities have a low correlation. So the IFCNN-SUM and CVT achieve the best performance on the metric $Q^{abf}$. Overall, our proposed fusion model achieves comparable performance for multi-modal medical images.

\section{Discussion}
As a tool to expand the measurement of optical lenses, multi-focus image fusion technology has the advantages of high efficiency and low cost, and in recent years, it has also been widely used in industrial, biological and medical fields. Specifically, it has been successfully applied to optical microscopes to enhance the imaging quality of microscopic views of microorganisms\cite{2005On} and to collect tiny biological samples\cite{2009Multifocus}. In medical-related fields, multi-focus image fusion technology is used to fuse microscopic cervical cell images for disease diagnosis\cite{2019Efficient}, and it is also used to accurately acquire Pap smear images\cite{2018Region}. In industry, multi-focus image fusion can be applied to structural measurements of nonwoven fabrics\cite{2019Structural} and can be equipped to microscope systems to improve the quality of the acquired printed circuit board (PCB) images\cite{pcb}. In general, multi-focus image fusion technology is relevant to our real life and therefore the proposed fusion framework is applicable for practical applications. However, there are some limitations to the implementation of multi-focus image fusion techniques. The multi-focus image fusion technique is more suitable for fusing static aligned images, and the real-time performance of the fusion technique still has much potential for improvement. Based on this, in our future research work, we will further improve the multi-focus image fusion algorithm from two aspects, namely, improving the processing speed of the algorithm and the implementation of real-time image alignment. Of course, this is also a reasonable discussion based on the problems of multi-focus image fusion technology in practical applications, and we hope to verify the feasibility of the improvement direction in the subsequent work.
\section{Conclusion}
In this paper, we propose a novel and effective image fusion model which can be trained in an end-to-end manner with supervised learning strategy. It avoids generating the intermediate decision map and utilizing post-process procedures to refine decision map for image fusion. Besides, to ﬁnely train the proposed fusion model, we have generated a large-scale multi-focus image dataset with ground-truth fusion images, and designed a novel fusion strategy based on the unity fusion attention, which can preserve more image information in the process of feature fusion. Specifically, the salient image features are firstly extracted from multiple input images. Then the proposed unity fusion attention is utilized to fuse these extracted convolutional features. Finally, the fused image features are fed into the cascaded convolutional blocks to reconstruct the fusion image. Extensive experiments are conducted to compare with other state-of-art fusion algorithms, which demonstrates that our proposed fusion approach can generate high-quality fusion images. In addition, the qualitative evaluation and quantitative evaluation performed on expanded experiments also indicated that our proposed fusion framework is effective in other image fusion tasks. 
\ifCLASSOPTIONcaptionsoff
  \newpage
\fi

\bibliographystyle{ieeetr}           
\bibliography{ref}

\begin{thebibliography}{10}

\bibitem{vishwakarma2018image}
A.~Vishwakarma and M.~K. Bhuyan, ``Image fusion using adjustable non-subsampled
  shearlet transform,'' {\em IEEE Transactions on Instrumentation and
  Measurement}, vol.~68, no.~9, pp.~3367--3378, 2018.

\bibitem{9345717}
H.~{Xu}, X.~{Wang}, and J.~{Ma}, ``Drf: Disentangled representation for visible
  and infrared image fusion,'' {\em IEEE Transactions on Instrumentation and
  Measurement}, vol.~70, pp.~1--13, 2021.

\bibitem{9187663}
L.~{Jian}, X.~{Yang}, Z.~{Liu}, G.~{Jeon}, M.~{Gao}, and D.~{Chisholm},
  ``Sedrfuse: A symmetric encoder–decoder with residual block network for
  infrared and visible image fusion,'' {\em IEEE Transactions on
  Instrumentation and Measurement}, vol.~70, pp.~1--15, 2021.

\bibitem{9305718}
Y.~{Yang}, S.~{Cao}, S.~{Huang}, and W.~{Wan}, ``Multimodal medical image
  fusion based on weighted local energy matching measurement and improved
  spatial frequency,'' {\em IEEE Transactions on Instrumentation and
  Measurement}, vol.~70, pp.~1--16, 2021.

\bibitem{8385209}
M.~{Yin}, X.~{Liu}, Y.~{Liu}, and X.~{Chen}, ``Medical image fusion with
  parameter-adaptive pulse coupled neural network in nonsubsampled shearlet
  transform domain,'' {\em IEEE Transactions on Instrumentation and
  Measurement}, vol.~68, no.~1, pp.~49--64, 2019.

\bibitem{2019Efficient}
Y.~Liang, Y.~Mao, Z.~Tang, M.~Yan, Y.~Zhao, and J.~Liu, ``Efficient
  misalignment-robust multi-focus microscopical images fusion,'' {\em Signal
  Processing}, vol.~161, no.~AUG., pp.~111--123, 2019.

\bibitem{2019Structural}
Yang, Chen, Na, Deng, Bin-Jie, Xin, Wen-Yu, Xing, and Zheng-Ye, ``Structural
  characterization and measurement of nonwoven fabrics based on multi-focus
  image fusion,'' {\em Measurement}, vol.~141, p.~356–363, 2019.

\bibitem{2018Region}
S.~Tello-Mijares and J.~Bescós, ``Region-based multifocus image fusion for the
  precise acquisition of pap smear images,'' {\em Journal of Biomedical
  Optics}, vol.~23, no.~5, p.~1, 2018.

\bibitem{stathaki2011image}
T.~Stathaki, {\em Image fusion: algorithms and applications}.
\newblock Elsevier, 2011.

\bibitem{toet1989morphological}
A.~Toet, ``A morphological pyramidal image decomposition,'' {\em Pattern
  Recognition Letters}, vol.~9, no.~4, pp.~255--261, 1989.

\bibitem{li1995multisensor}
H.~Li, B.~Manjunath, and S.~K. Mitra, ``Multisensor image fusion using the
  wavelet transform,'' {\em Graphical Models and Image Processing}, vol.~57,
  no.~3, pp.~235--245, 1995.

\bibitem{2018A}
S.~Aymaz and C.~Kse, ``A novel image decomposition-based hybrid technique with
  super-resolution method for multi-focus image fusion,'' {\em Information
  Fusion}, vol.~45, pp.~113--127, 2018.

\bibitem{lewis2007pixel}
J.~J. Lewis, R.~J. O’Callaghan, S.~G. Nikolov, D.~R. Bull, and
  N.~Canagarajah, ``Pixel- and region-based image fusion with complex
  wavelets,'' {\em Information Fusion}, vol.~8, no.~2, pp.~119--130, 2007.

\bibitem{zhang2009multifocus}
Q.~Zhang and B.-l. Guo, ``Multifocus image fusion using the nonsubsampled
  contourlet transform,'' {\em Signal Processing}, vol.~89, no.~7,
  pp.~1334--1346, 2009.

\bibitem{bai2015quadtree}
X.~Bai, Y.~Zhang, F.~Zhou, and B.~Xue, ``Quadtree-based multi-focus image
  fusion using a weighted focus-measure,'' {\em Information Fusion}, vol.~22,
  pp.~105--118, 2015.

\bibitem{li2013image}
S.~Li, X.~Kang, and J.~Hu, ``Image fusion with guided filtering,'' {\em IEEE
  Transactions on Image processing}, vol.~22, no.~7, pp.~2864--2875, 2013.

\bibitem{li2008multifocus}
S.~Li and B.~Yang, ``Multifocus image fusion using region segmentation and
  spatial frequency,'' {\em Image and Vision Computing}, vol.~26, no.~7,
  pp.~971--979, 2008.

\bibitem{zhou2014multi}
Z.~Zhou, S.~Li, and B.~Wang, ``Multi-scale weighted gradient-based fusion for
  multi-focus images,'' {\em Information Fusion}, vol.~20, pp.~60--72, 2014.

\bibitem{liu2015multi}
Y.~Liu, S.~Liu, and Z.~Wang, ``Multi-focus image fusion with dense sift,'' {\em
  Information Fusion}, vol.~23, pp.~139--155, 2015.

\bibitem{liu2017multi}
Y.~Liu, X.~Chen, H.~Peng, and Z.~Wang, ``Multi-focus image fusion with a deep
  convolutional neural network,'' {\em Information Fusion}, vol.~36,
  pp.~191--207, 2017.

\bibitem{tang2018pixel}
H.~Tang, B.~Xiao, W.~Li, and G.~Wang, ``Pixel convolutional neural network for
  multi-focus image fusion,'' {\em Information Sciences}, vol.~433,
  pp.~125--141, 2018.

\bibitem{ma2020sesf}
B.~Ma, Y.~Zhu, X.~Yin, X.~Ban, H.~Huang, and M.~Mukeshimana, ``Sesf-fuse: An
  unsupervised deep model for multi-focus image fusion,'' {\em Neural Computing
  and Applications}, pp.~1--12, 2020.
\newblock doi:10.1007/s00521-020-05358-9.

\bibitem{prabhakar2017deepfuse}
K.~R. {Prabhakar}, V.~S. {Srikar}, and R.~V. {Babu}, ``Deepfuse: A deep
  unsupervised approach for exposure fusion with extreme exposure image
  pairs,'' in {\em 2017 IEEE International Conference on Computer Vision
  (ICCV)}, pp.~4724--4732, 2017.

\bibitem{li2018densefuse}
H.~Li and X.~Wu, ``Densefuse: A fusion approach to infrared and visible
  images,'' {\em IEEE Transactions on Image Processing}, vol.~28, no.~5,
  pp.~2614--2623, 2018.

\bibitem{zhang2020ifcnn}
Y.~Zhang, Y.~Liu, P.~Sun, H.~Yan, X.~Zhao, and L.~Zhang, ``Ifcnn: A general
  image fusion framework based on convolutional neural network,'' {\em
  Information Fusion}, vol.~54, pp.~99--118, 2020.

\bibitem{nencini2007remote}
F.~Nencini, A.~Garzelli, S.~Baronti, and L.~Alparone, ``Remote sensing image
  fusion using the curvelet transform,'' {\em Information Fusion}, vol.~8,
  no.~2, pp.~143--156, 2007.

\bibitem{chen2017robust}
Y.~Chen, J.~Guan, and W.-K. Cham, ``Robust multi-focus image fusion using edge
  model and multi-matting,'' {\em IEEE Transactions on Image Processing},
  vol.~27, no.~3, pp.~1526--1541, 2017.

\bibitem{9020016}
J.~Li, X.~Guo, G.~Lu, B.~Zhang, and D.~Zhang, ``Drpl: Deep regression pair
  learning for multi-focus image fusion,'' {\em IEEE Transactions on Image
  Processing}, vol.~PP, no.~99, pp.~1--1, 2020.

\bibitem{ZHANG202140}
H.~Zhang, Z.~Le, Z.~Shao, H.~Xu, and J.~Ma, ``Mff-gan: An unsupervised
  generative adversarial network with adaptive and gradient joint constraints
  for multi-focus image fusion,'' {\em Information Fusion}, vol.~66,
  pp.~40--53, 2021.

\bibitem{2020Brain}
Z.~Ding, D.~Zhou, R.~Nie, R.~Hou, and Y.~Liu, ``Brain medical image fusion
  based on dual-branch cnns in nsst domain,'' {\em BioMed Research
  International}, vol.~2020, pp.~1--15, 2020.

\bibitem{guo2020deepanf}
Y.~Guo, D.~Zhou, R.~Nie, X.~Ruan, and W.~Li, ``Deepanf: A deep attentive neural
  framework with distributed representation for chromatin accessibility
  prediction,'' {\em Neurocomputing}, vol.~379, pp.~305--318, 2020.

\bibitem{hu2018squeeze}
J.~Hu, L.~Shen, and G.~Sun, ``Squeeze-and-excitation networks,'' in {\em
  Proceedings of the IEEE Conference on Computer Vision and Pattern
  Recognition}, pp.~7132--7141, 2018.

\bibitem{9216075}
J.~{Li}, H.~{Huo}, C.~{Li}, R.~{Wang}, C.~{Sui}, and Z.~{Liu}, ``Multigrained
  attention network for infrared and visible image fusion,'' {\em IEEE
  Transactions on Instrumentation and Measurement}, vol.~70, pp.~1--12, 2021.

\bibitem{9127964}
H.~Li, X.~J. Wu, and T.~Durrani, ``Nestfuse: An infrared and visible image
  fusion architecture based on nest connection and spatial/channel attention
  models,'' {\em IEEE Transactions on Instrumentation and Measurement},
  vol.~PP, no.~99, pp.~1--1, 2020.

\bibitem{9242278}
B.~{Xiao}, B.~{Xu}, X.~{Bi}, and W.~{Li}, ``Global-feature encoding u-net
  (geu-net) for multi-focus image fusion,'' {\em IEEE Transactions on Image
  Processing}, vol.~30, pp.~163--175, 2021.

\bibitem{woo2018cbam}
S.~Woo, J.~Park, J.~Lee, and I.~So~Kweon, ``Cbam: Convolutional block attention
  module,'' in {\em Proceedings of the European Conference on Computer Vision
  (ECCV)}, pp.~3--19, 2018.

\bibitem{7797130}
H.~{Zhao}, O.~{Gallo}, I.~{Frosio}, and J.~{Kautz}, ``Loss functions for image
  restoration with neural networks,'' {\em IEEE Transactions on Computational
  Imaging}, vol.~3, no.~1, pp.~47--57, 2017.

\bibitem{lin2014microsoft}
T.~Lin, M.~Maire, S.~Belongie, J.~Hays, P.~Perona, D.~Ramanan, P.~Doll{\'a}r,
  and C.~L. Zitnick, ``Microsoft coco: Common objects in context,'' in {\em
  European Conference on Computer Vision}, pp.~740--755, Springer, 2014.

\bibitem{wang2004image}
Z.~Wang, A.~C. Bovik, H.~R. Sheikh, and E.~P. Simoncelli, ``Image quality
  assessment: from error visibility to structural similarity,'' {\em IEEE
  Transactions on Image Processing}, vol.~13, no.~4, pp.~600--612, 2004.

\bibitem{wang2017learning}
L.~Wang, H.~Lu, Y.~Wang, M.~Feng, D.~Wang, B.~Yin, and X.~Ruan, ``Learning to
  detect salient objects with image-level supervision,'' in {\em Proceedings of
  the IEEE Conference on Computer Vision and Pattern Recognition},
  pp.~136--145, 2017.

\bibitem{kingma2014adam}
D.~P. Kingma and J.~Ba, ``Adam: A method for stochastic optimization,'' {\em
  arXiv preprint arXiv:1412.6980}, 2014.

\bibitem{yang2009multifocus}
B.~Yang and S.~Li, ``Multifocus image fusion and restoration with sparse
  representation,'' {\em IEEE Transactions on Instrumentation and Measurement},
  vol.~59, no.~4, pp.~884--892, 2009.

\bibitem{han2013new}
Y.~Han, Y.~Cai, Y.~Cao, and X.~Xu, ``A new image fusion performance metric
  based on visual information fidelity,'' {\em Information Fusion}, vol.~14,
  no.~2, pp.~127--135, 2013.

\bibitem{liu2005image}
G.~Liu, W.~Chen, and W.~Ling, ``An image fusion method based on directional
  contrast and area-based standard deviation,'' in {\em Electronic Imaging and
  Multimedia Technology IV}, vol.~5637, pp.~50--56, International Society for
  Optics and Photonics, 2005.

\bibitem{song2007fusion}
H.~Song, S.~Yu, L.~Song, and X.~Yang, ``Fusion of multispectral and
  panchromatic satellite images based on contourlet transform and local average
  gradient,'' {\em Optical Engineering}, vol.~46, no.~2, p.~020502, 2007.

\bibitem{emmerich2007gradient}
M.~Emmerich, A.~Deutz, and N.~Beume, ``Gradient-based/evolutionary relay hybrid
  for computing pareto front approximations maximizing the s-metric,'' in {\em
  International Workshop on Hybrid Metaheuristics}, pp.~140--156, Springer,
  2007.

\bibitem{wax1977improved}
M.~Wax and J.~Ziv, ``Improved bounds on the local mean-square error and the
  bias of parameter estimators (corresp.),'' {\em IEEE Transactions on
  Information Theory}, vol.~23, no.~4, pp.~529--530, 1977.

\bibitem{chen2009new}
Y.~Chen and R.~S. Blum, ``A new automated quality assessment algorithm for
  image fusion,'' {\em Image and Vision Computing}, vol.~27, no.~10,
  pp.~1421--1432, 2009.

\bibitem{wang2008novel}
P.~Wang and B.~Liu, ``A novel image fusion metric based on multi-scale
  analysis,'' in {\em 2008 9th International Conference on Signal Processing},
  pp.~965--968, IEEE, 2008.

\bibitem{nejati2015multi}
M.~Nejati, S.~Samavi, and S.~Shirani, ``Multi-focus image fusion using
  dictionary-based sparse representation,'' {\em Information Fusion}, vol.~25,
  pp.~72--84, 2015.

\bibitem{ma2016infrared}
J.~Ma, C.~Chen, C.~Li, and J.~Huang, ``Infrared and visible image fusion via
  gradient transfer and total variation minimization,'' {\em Information
  Fusion}, vol.~31, pp.~100--109, 2016.

\bibitem{zhou2016perceptual}
Z.~Zhou, B.~Wang, S.~Li, and M.~Dong, ``Perceptual fusion of infrared and
  visible images through a hybrid multi-scale decomposition with gaussian and
  bilateral filters,'' {\em Information Fusion}, vol.~30, pp.~15--26, 2016.

\bibitem{hou2020vif}
R.~Hou, D.~Zhou, R.~Nie, D.~Liu, L.~Xiong, Y.~Guo, and C.~Yu, ``Vif-net: an
  unsupervised framework for infrared and visible image fusion,'' {\em IEEE
  Transactions on Computational Imaging}, vol.~6, pp.~640--651, 2020.

\bibitem{liu2017medical}
Y.~Liu, X.~Chen, J.~Cheng, and H.~Peng, ``A medical image fusion method based
  on convolutional neural networks,'' in {\em 2017 20th International
  Conference on Information Fusion (Fusion)}, pp.~1--7, IEEE, 2017.

\bibitem{singh2019multimodal}
S.~Singh and R.~Anand, ``Multimodal medical image fusion using hybrid layer
  decomposition with cnn-based feature mapping and structural clustering,''
  {\em IEEE Transactions on Instrumentation and Measurement}, vol.~69, no.~6,
  pp.~3855--3865, 2019.

\bibitem{li2020laplacian}
X.~Li, X.~Guo, P.~Han, X.~Wang, H.~Li, and T.~Luo, ``Laplacian re-decomposition
  for multimodal medical image fusion,'' {\em IEEE Transactions on
  Instrumentation and Measurement}, vol.~69, no.~9, pp.~6880--6890, 2020.

\bibitem{2005On}
S.~Gabarda and G.~Cristobal, ``On the use of a joint spatial-frequency
  representation for the fusion of multi-focus images,'' {\em Pattern
  Recognition Letters}, vol.~26, no.~16, pp.~2572--2578, 2005.

\bibitem{2009Multifocus}
R.~Redondo, F.~Sroubek, S.~Fischer, and G.~Cristóbal, ``Multifocus image
  fusion using the log-gabor transform and a multisize windows technique,''
  {\em Information Fusion}, vol.~10, no.~2, pp.~163--171, 2009.

\bibitem{pcb}
V.~R. L.~Juocas, R.~D. R.~Maskeliunas, and M.~Wozniak, ``Multi-focus- ing
  algorithm for microscopy imagery in assembly line using low-cost camera,''
  {\em Journal of Biomedical Optics}, vol.~102, no.~9-12, p.~3217–3227, 2019.

\end{thebibliography}
  
\begin{IEEEbiography}[{\includegraphics[width=1in,height=1.25in,clip,keepaspectratio]{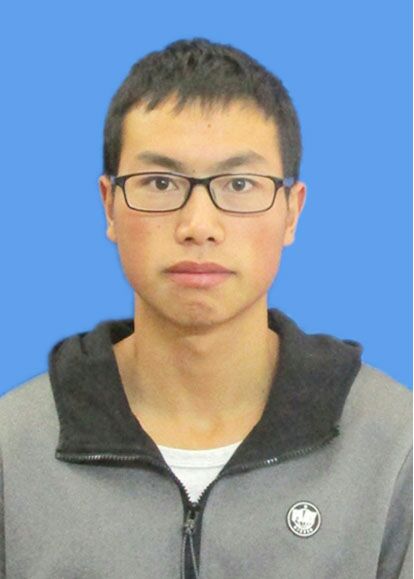}}]{Yongsheng Zang}
	received the B.E. degree in Electrical Engineering and Automation from Yunnan University, Kunming, China, in 2019, and he is currently pursuing the M.S. degree in Yunnan University.  His current research interests include image fusion, image super-resolution and deep learning.
\end{IEEEbiography}
\vspace{5mm}
\begin{IEEEbiography}[{\includegraphics[width=1in,height=1.25in,clip,keepaspectratio]{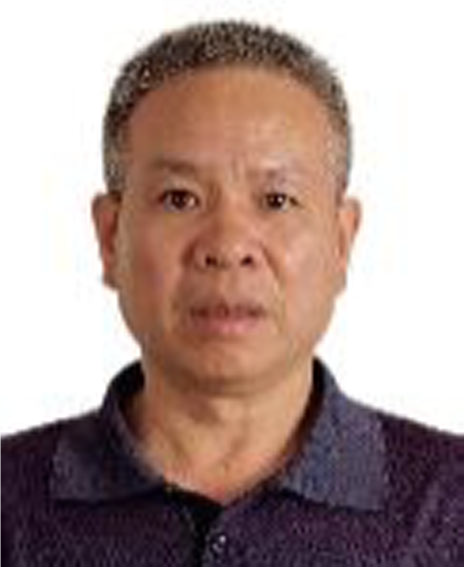}}]{Dongming Zhou}
	received the B.S. and M.S. degrees in Automatic Control Engineering from the Hua Zhong University of Science and Technology (HUST). Wuhan. China, in 1985 and 1988. respectively, and the Ph.D. degree in Electronic Circuit and System from Fudan University, Shanghai, China.in 2004. He is currently a Professor at Yunnan University. Kunming, China. His current research interests include neural network theory.
\end{IEEEbiography}
\vspace{5mm}
\begin{IEEEbiography}[{\includegraphics[width=1in,height=1.25in,clip,keepaspectratio]{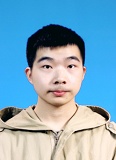}}]{Changcheng Wang}
	received the B.E. degree in School of Computer and Communication Engineering from Changsha University of Science and Technology, Changsha, China, in 2018. He is currently pursuing the M.S degree with the School of Information Science and Engineering, Yunnan University, Kunming, China. His interests include neural networks, image processing and image fusion. 
\end{IEEEbiography}
\vspace{2mm}
\begin{IEEEbiography}[{\includegraphics[width=1in,height=1.25in,clip,keepaspectratio]{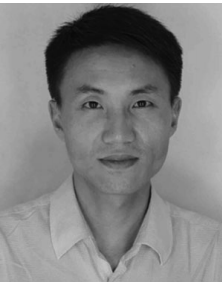}}]{Rencan Nie}	
	received the Ph.D. degree in Tele Communication Engineering from Yunnan University, Kunming, China, in 2014.He is currently an Associate Professor with Yunnan University. His current research interests include image processing and machine learning.
\end{IEEEbiography}
\vspace{5mm}
\begin{IEEEbiography}[{\includegraphics[width=1in,height=1.25in,clip,keepaspectratio]{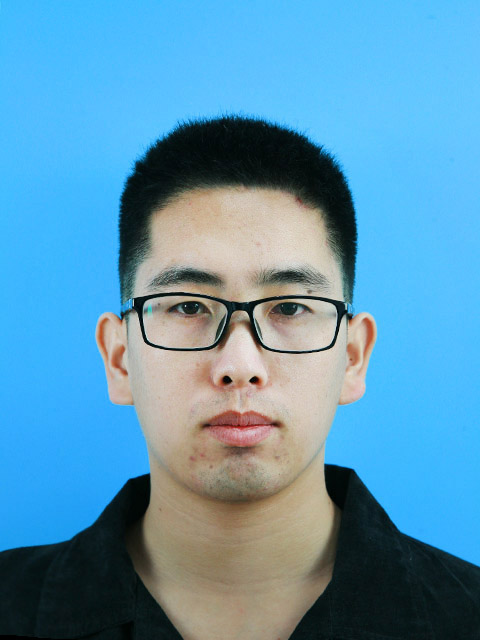}}]{Yanbu Guo}received his B.S. degree in College of Mathematics and Information Science from Zhengzhou University of Light Industry, Zhengzhou, China, in 2016. He is currently pursuing a Ph.D. degree with the School of Information Science and Engineering, Yunnan University, Kunming, China. His research interests include neural networks, computational intelligence, biomedical and health informatics.
\end{IEEEbiography}
\end{document}